%% file: main.tex
\documentclass{vislang_lab_2026}

\usepackage{mathtools}
\usepackage[numbers,sort&compress]{natbib}
\usepackage{subcaption}
\usepackage{multirow}
\usepackage{float}
\usepackage{xspace}
\usepackage[capitalize,noabbrev]{cleveref}

\usepackage{silence}
\definecolor{darkblue}{rgb}{0,0,0.5}

\newtcolorbox{prompt}[1]{
  enhanced,
  left=2mm,
  right=2mm,
  top=1mm,
  bottom=1mm,
  boxsep=0mm,
  boxrule=0.4pt,
  rounded corners,
  title=#1,
  fontupper=\scriptsize\linespread{0.88}\selectfont,
}
\newenvironment{lmttfont}{\fontfamily{lmtt}\selectfont}{\par}

\newcommand{\custompara}[1]{{\vspace{1mm}\noindent\textbf{#1}\xspace}}
\newcommand{\grayfont}{\color{gray}}
\newcommand{\method}{SoftmaxGRPO}

\title{\method{}: Learning to Reason using Softmax Advantage Group Estimation}

\author[1]{Jefferson Hernandez}
\author[1]{Jaywon Koo}
\author[1]{Zilin Xiao}
\author[1]{Chen Wei}
\author[1]{Vicente Ordonez}
\affil[1]{Rice University}
\correspondingauthor={{}\href{mailto:jefehern@rice.edu}{jefehern@rice.edu}}

\begin{document}

\maketitle

\input{sections/0_abstract}
\input{sections/1_intro}
\input{sections/2_related}
\input{sections/3_method}
\input{sections/4_experiments}
\input{sections/5_conclusions}

{
  \small
  \bibliographystyle{plainnat}
  \bibliography{sections/6_references}
}

\input{sections/7_appendix}

\end{document}

%% file: sections/0_abstract.tex
\begin{abstract}

Group-based reinforcement learning objectives such as GRPO can allocate learning signal poorly across prompt difficulty: under binary rewards, group normalization induces a divergent weighting on easy prompts. We introduce Softmax Advantage Group Estimation (\method{}), a drop-in alternative that replaces z-score-normalized group advantages with temperature-scaled softmax advantages, keeping weights bounded regardless of prompt difficulty. For binary rewards, we derive the exact finite-group population objective and identify MaxRL as its low-temperature limit. For bounded scalar rewards, we show that the large-group update exactly optimizes a log-moment-generating-function objective, while a universal finite-group scalar objective cannot exist without additional assumptions on the reward distribution. Empirically, \method{} reallocates measured gradient budget away from near-solved prompts and consistently improves over GRPO under identical rewards. It reaches 51.8\% on DeepMath with verifiable rewards and improves a 1.5B instruction-tuned model from 35.0\% to 68.0\% on Poetry using only lightweight text-similarity rewards.

\end{abstract}

%% file: sections/1_intro.tex
\section{Introduction}
\label{sec:intro}

Group-based reinforcement learning objectives such as GRPO have become standard for post-training~\citep{shao2024deepseekmath,guo2025rlvr,deepseek_analysis}: sample $M$ rollouts per prompt, normalize within-group rewards, and optimize a PPO-style surrogate. The normalization choice is not cosmetic. Under binary correctness rewards, groupwise objectives are best understood as optimizing different monotone transforms of pass probability, each inducing a distinct prompt-difficulty weighting~\citep{davis2025objective,tajwar2026maxrl}. GRPO normalization induces a weighting that diverges on easy prompts, over-concentrating gradient signal on problems the model already solves reliably. Practical variants---Dr.GRPO, DAPO, CISPO, DPPO~\citep{qi2026rethinking,yu2025dapo,liu2025understanding,chen2025minimax}---address symptoms of this imbalance without replacing the underlying objective geometry.

This pathology is especially damaging outside the narrow regime of tasks with cheap automatic verifiers. Many settings of practical interest---summarization, open-ended question answering, creative generation---supply only weak answer-level signals such as string-overlap scores (ROUGE, BLEU) against reference outputs. Weak rewards are noisy and sparse; a poorly shaped objective that wastes gradient budget on already-easy prompts compounds this problem, leaving even less signal where learning is most needed.

\begin{figure}[t!]
    \centering
    \begin{subfigure}[b]{0.48\textwidth}
        \centering
        \includegraphics[width=\textwidth]{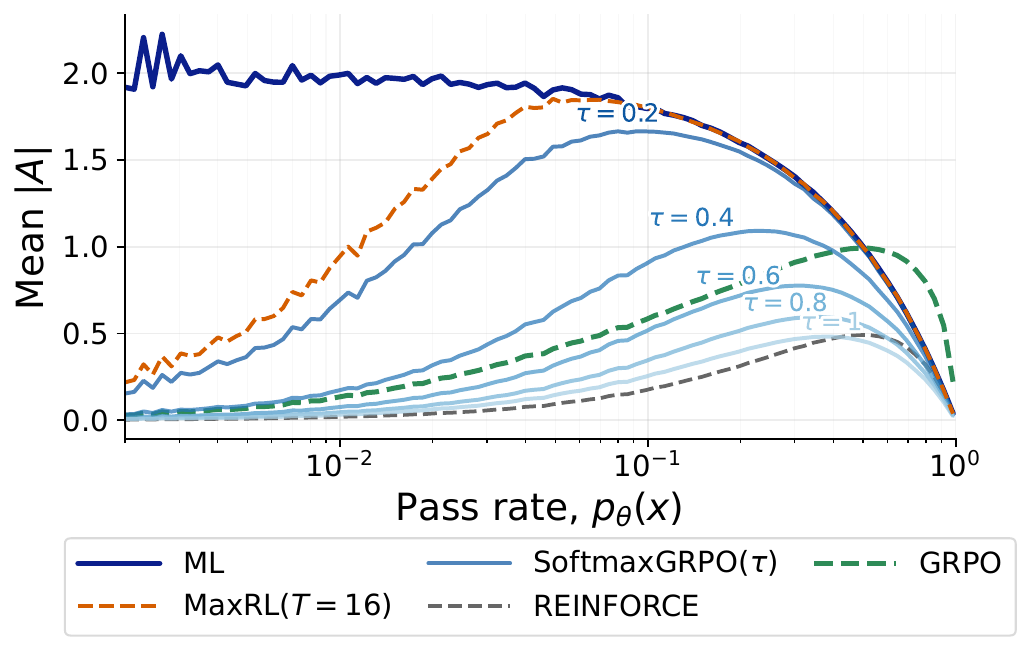}
        \caption{Mean absolute advantage for ML, GRPO, REINFORCE, MaxRL, and \method{}($\tau$) across pass rates, estimated by Monte Carlo over groups of $M$ binary rollouts.}
        \label{fig:adv_mag}
    \end{subfigure}
    \hfill
    \begin{subfigure}[b]{0.48\textwidth}
        \centering
        \includegraphics[width=\textwidth]{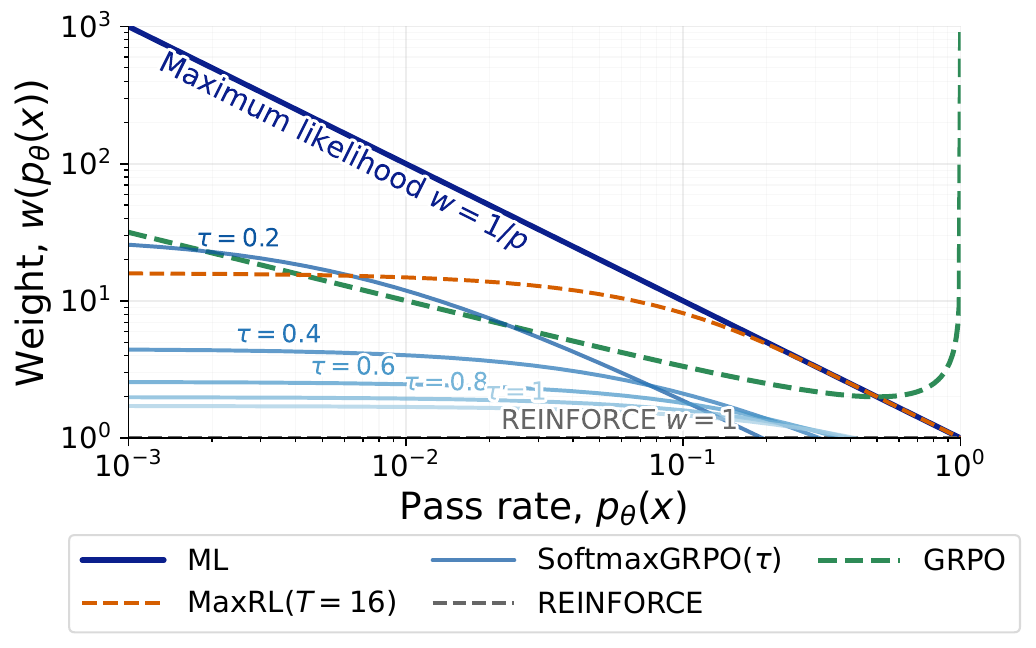}
        \caption{Population-level prompt-weighting functions $w(p)$ as a function of pass rate $p$ for ML, GRPO, REINFORCE, MaxRL, and \method{}($\tau$).}
        \label{fig:weight_func}
    \end{subfigure}

    \caption{\textbf{\method{} defines a smooth objective family over prompt difficulty.} (a) Adjusting $\tau$ changes the gradient signal magnitude, moving from REINFORCE-like toward MaxRL-like behavior at finite $M$. (b) The induced population weights show how \method{} reallocates learning signal as pass rate changes. The joint low-temperature, large-group limit approaches ML weighting, while \method{} avoids GRPO's easy-prompt blow-up.}
    \label{fig:combined_figures}
\end{figure}

We introduce \textbf{Softmax Advantage Group Estimation (\method{})}, a one-line replacement for GRPO: given $M$ rollouts with rewards $\{r_i\}$, form within-group weights $w_i \propto \exp(r_i/\tau)$ and centered advantages $A_i = Mw_i - 1$. For binary rewards in the unclipped on-policy regime, \method{} induces an exact finite-$M$ objective $h_{M,\tau}(p)$ with bounded prompt weighting; Figure~\ref{fig:combined_figures} illustrates how $\tau$ moves \method{} between REINFORCE-like and MaxRL-like behavior, approaching maximum-likelihood weighting only in the joint low-temperature, large-group limit. For bounded scalar rewards, the large-group update exactly optimizes the log moment-generating function of reward. This result is also sharp: with three or more reward levels, the finite-group update is generally non-conservative, so no universal scalar analogue of $h_{M,\tau}$ exists without additional assumptions. At finite $M$, the weights retain the standard RAML/MPO-style exponential-tilting interpretation~\citep{norouzi2016raml,abdolmaleki2018mpo}; in experiments, we optimize them with PPO clipping and reference-model KL regularization.

\noindent Our contributions are:
\begin{itemize}
    \item We introduce \method{}, a one-line drop-in replacement for GRPO that substitutes temperature-scaled softmax advantages for z-score group advantages, keeping weights bounded at all pass rates.
    \item We derive the exact finite-$M$ binary-reward objective, establish its MaxRL limit, and prove an exact large-group objective for bounded scalar rewards. We also show why a universal finite-$M$ scalar objective generally cannot exist beyond binary rewards.
    \item Empirically, \method{} reallocates gradient budget away from near-solved prompts, outperforms GRPO under identical weak rewards, and performs strongly across both verifiable and non-verifiable tasks.
\end{itemize}

%% file: sections/2_related.tex
\section{Related Work}
\label{sec:related}

\custompara{Reasoning RL and group-based policy objectives.}
Reinforcement learning with verifiable rewards (RLVR) has driven strong gains on mathematical reasoning and code generation by optimizing outcome signals from automatic checkers at scale~\citep{shao2024deepseekmath, guo2025rlvr, jaech2024openai}. More recently, this type of model post-training has also proven useful in multimodal reasoning~\citep{vlrethinker,openvlthinker,thinklitevl,xiao2026proxythinker,he2026referringexpressionsscenariocomprehension,xia2026sportrbenchmarkmultimodallarge}.  Within this paradigm, group-based objectives such as GRPO have become standard, normalizing within-group rewards before forming policy-gradient targets. Several practical variants address instabilities arising from this normalization through adaptive clipping, filtering, and reweighting schemes~\citep{qi2026rethinking, yu2025dapo, liu2025understanding, chen2025minimax}. Rather than patching the normalization, \method{} replaces it with a temperature-scaled softmax that admits an exact population-level analysis and naturally interpolates between distinct optimization regimes.

\custompara{RAML, softmax policy gradient, and exponentiated-reward methods.}
Exponentiated-reward weighting itself is well established. Reward-Augmented Maximum Likelihood (RAML) forms reward-shaped maximum-likelihood targets~\citep{norouzi2016raml}; Optimal Completion Distillation extends related ideas to per-prefix targets~\citep{sabour2018ocd}; and softmax policy gradient and MPO use exponentiated-reward or advantage targets for policy improvement~\citep{ding2017softmaxpg,abdolmaleki2018mpo}. Our contribution is not the softmax construction, but its group-level geometry: the exact finite-$M$ prompt-weighting objective under binary rewards, its MaxRL limit, the large-group scalar-reward objective, and the finite-group obstruction beyond binary rewards.

\custompara{Weak-rewards and non-verifiable training.}
Extending RLVR beyond verifiable domains requires either domain-specific reward models or tolerance for weak, noisy signals such as string-overlap metrics against reference outputs~\citep{ma2025general, guo2025rlvr}. When only the final answer is scored, the training signal provides little guidance about which parts of a long generation should change and can incentivize rationales weakly tied to the actual computation~\citep{huang2025misalignment}; process-level constraints partially address this but inherit the same fundamental sparsity~\citep{tu2025autothink}. Self-supervised objectives derived from unlabeled text suggest appealing scaling properties but are typically coupled to pre-training rather than post-training fine-tuning~\citep{hatamizadeh2025rlp}. \method{} targets this harder post-training regime directly, using standard gold-answer supervision with on-policy sampling to reshape objective geometry, rather than requiring step-by-step annotations or a separate reward model.

%% file: sections/3_method.tex
\section{\method{}: Softmax Advantage Group Estimation}
\label{sec:sage}

\method{} is a one-line replacement for GRPO-style z-score normalization in PPO-based reasoning post-training~\citep{schulman2017ppo,shao2024deepseekmath}: given a group of $M$ rollouts, replace standardized rewards with a softmax over rewards. The cleanest population picture comes from binary rewards, where this 
choice induces a prompt-weighting family over pass probability $p$. We make that weight function the organizing principle of this section.

\subsection{Update rule}
\label{subsec:sage_update}

For an input $x$, let $\mathcal G(x)=\{z_i\}_{i=1}^M$ be a sampled rollout group, decode $y_i=\mathrm{Dec}(z_i)$, and compute rewards $R_i := R(x,y_i)$. \method{} forms
\begin{equation}
w_i
=
\frac{\exp(R_i/\tau)}{\sum_{j=1}^M \exp(R_j/\tau)},
\qquad
A_i
=
M w_i - 1.
\label{eq:sage_weights}
\end{equation}
Since $\sum_{i=1}^M A_i = 0$, \method{} is invariant to additive reward shifts within the group and produces no update when all sampled rollouts receive the same reward. The underlying unclipped group objective is
\begin{equation}
\mathcal J_{\mathrm{\method{}}}^{\mathrm{uc}}(\theta;x,\mathcal G)
=
\frac{1}{M}\sum_{i=1}^M A_i \log \pi_\theta(z_i\mid x).
\label{eq:sage_uc}
\end{equation}

\subsection{\method{}  under binary rewards}
\label{subsec:sage_weight_function}

We now specialize in binary correctness rewards and in the unclipped on-policy regime. For binary rewards, the expected \method{} update on a prompt with pass probability $p$ takes the form $\omega(p)\,\nabla_\theta p_\theta(x)$. In large groups, that weight can be read off directly from the softmax.

Let $p := p_\theta(x)$ and $c := e^{1/\tau}$. In a large group, a fraction $p$ of rollouts are correct and carry unnormalized softmax mass $c$, while a fraction $1-p$ are incorrect and carry mass $1$. The total within-group mass is therefore $1-p+pc$, and the success--failure gap induced by \method{} is
\begin{equation}
\omega_\tau(p)
\approx
\frac{c-1}{1-p+pc}
=
\frac{e^{1/\tau}-1}{1-p+p e^{1/\tau}}.
\label{eq:sage_weight_mf}
\end{equation}
This is the main theoretical insight: \method{} defines a temperature-controlled family of prompt-weighting functions. Large $\tau$ makes $\omega_\tau(p)$ nearly constant, recovering REINFORCE-like weighting; small $\tau$ shifts mass toward hard prompts, approaching $1/p$ and thus maximum-likelihood-style weighting in the joint low-temperature, large-group limit. Unlike GRPO, whose binary-reward weighting scales as $[p(1-p)]^{-1/2}$~\citep{davis2025objective,tajwar2026maxrl}, \method{} remains finite as $p\to 1$, so it does not overemphasize already-solved prompts.

As throughout this comparison, only the shape of $\omega(p)$ matters; positive $p$-independent rescalings can be absorbed into the step size in the unclipped population picture.

\begin{table*}[t]
\centering
\small
\setlength{\tabcolsep}{6pt}
\renewcommand{\arraystretch}{1.12}
\begin{tabular}{l c p{8.0cm}}
\toprule
\textbf{Objective} & \textbf{Weight $\omega(p)$} & \textbf{Geometry} \\
\midrule
REINFORCE
& $1$
& Uniform weighting over prompt difficulty. \\

ML
& $\frac{1}{p}$
& Train-on-successes geometry; strongly emphasizes hard prompts. \\

GRPO
& $\frac{1}{\sqrt{p(1-p)}}$
& Blows up on both very hard and already-easy prompts. \\

MaxRL$(T)$
& $\frac{1-(1-p)^T}{p}$
& Truncated ML weighting; caps the hard-prompt blow-up for finite $T$. \\

\method{}
& $\frac{e^{1/\tau}-1}{1-p+p e^{1/\tau}}$
& REINFORCE-to-MaxRL interpolation at finite $M$; ML-like in the joint low-temperature, large-group limit; finite at $p\to1$. \\
\bottomrule
\end{tabular}
\caption{\textbf{Prompt-weighting view of binary-reward objectives.}
We show weight functions up to positive $p$-independent scaling. \method{} is shown in its large-$M$ form; the exact finite-$M$ weight is given in Eq.~\eqref{eq:sage_exact_weight}, and the low-temperature finite-$M$ limit in Eq.~\eqref{eq:sage_maxrl}.}
\label{tab:weight_view}
\end{table*}

\subsection{Limit behavior of \method{}}
\label{subsec:sage_exact_finiteM}

The mean-field form above is the main intuition. The exact finite-$M$ binary-reward objective has the same structure. Let
$S_i := \sum_{j\neq i} R_j$
be the number of other successful rollouts seen by sample $i$. Conditioned on $S_i=s$, a successful rollout and an unsuccessful rollout receive weights
\[
w^{(1)}_s
=
\frac{c}{M+(s+1)(c-1)},
\qquad
w^{(0)}_s
=
\frac{1}{M+s(c-1)}.
\]
Their centered-advantage gap is therefore
\[
\Delta_s^{(\tau)}
:=
M\bigl(w^{(1)}_s - w^{(0)}_s\bigr).
\]
Averaging over $S\sim \mathrm{Binomial}(M-1,p)$ yields the exact prompt-weighting function
\begin{equation}
\begin{split}
\mathbb E_{\mathcal G(x)\sim \pi_\theta(\cdot\mid x)}
\!\left[
\nabla_\theta \mathcal J_{\mathrm{\method{}}}^{\mathrm{uc}}(\theta;x,\mathcal G)
\right]
&=
\omega_{M,\tau}(p_\theta(x))\,\nabla_\theta p_\theta(x), \\
\omega_{M,\tau}(p)
&=
\mathbb E_{S\sim \mathrm{Binomial}(M-1,p)}
\!\left[\Delta_S^{(\tau)}\right].
\end{split}
\label{eq:sage_exact_weight}
\end{equation}
Thus \method{} optimizes a scalar transform $h_{M,\tau}(p)$ with derivative $h_{M,\tau}'(p)=\omega_{M,\tau}(p)$. The closed-form expression for $h_{M,\tau}$, together with the Bernstein-polynomial representation of $\omega_{M,\tau}$, is given in Appendix~\ref{app:sage_exact}.

Most importantly, the low-temperature finite-$M$ limit is exact:
\begin{equation}
\omega_{M,0}(p)
=
\frac{1-(1-p)^{M-1}}{p}.
\label{eq:sage_maxrl}
\end{equation}
This is exactly the MaxRL weighting with truncation $T=M-1$~\citep{tajwar2026maxrl}. \method{} is a smooth temperature-parameterized method that interpolates from REINFORCE to MaxRL as $\tau$ shrinks, and then to ML-like $1/p$ weighting as $M$ grows.

\subsection{General scalar rewards: objective and limitation}
\label{subsec:sage_general_rewards}

The binary result above is exact at finite $M$. A different exact objective emerges for bounded scalar rewards in the large-group limit. Define
\begin{equation}
Z_\tau(\theta;x)
:=
\mathbb E_{z\sim\pi_\theta(\cdot\mid x)}
\!\left[e^{R(x,z)/\tau}\right].
\label{eq:sage_partition}
\end{equation}
For an i.i.d.\ on-policy group, the softmax denominator concentrates around $MZ_\tau$. Applying the score identity to Eq.~\eqref{eq:sage_uc} therefore gives the exact limit
\begin{equation}
\begin{split}
\lim_{M\to\infty}
\mathbb E_{\mathcal G(x)}
\!\left[\nabla_\theta \mathcal J_{\mathrm{\method{}}}^{\mathrm{uc}}\right]
&=
\mathbb E_{z\sim\pi_\theta}
\!\left[
\left(\frac{e^{R(x,z)/\tau}}{Z_\tau(\theta;x)}-1\right)
\nabla_\theta\log\pi_\theta(z\mid x)
\right] \\
&=
\nabla_\theta\log Z_\tau(\theta;x).
\end{split}
\label{eq:sage_general_objective}
\end{equation}
Thus large-group \method{} optimizes the log moment-generating function of reward, or equivalently the exponential-utility objective up to a positive factor of $\tau$. For binary reward, $Z_\tau=1-p+p e^{1/\tau}$, and Eq.~\eqref{eq:sage_general_objective} recovers the large-$M$ weight in Eq.~\eqref{eq:sage_weight_mf} exactly.
Appendix~\ref{app:general_scalar} gives the full proof, finite-$M$ counterexample, and Gaussian-reward example; Appendix~\ref{app:noisy_binary} shows stability to small bounded noise around binary rewards.

This large-group result cannot generally be strengthened to a finite-$M$ scalar objective. Consider $M=2$ and three reward levels with probabilities $p_k$ and $t_k=e^{r_k/\tau}$. The expected coefficient for level $k$ is
$m_k(p)=2\sum_\ell p_\ell t_k/(t_k+t_\ell)$.
On the simplex $p_3=1-p_1-p_2$, a scalar potential would require the one-form $(m_1-m_3)\,dp_1+(m_2-m_3)\,dp_2$ to be closed. For $t=(1,2,4)$, however,
\begin{equation}
\frac{\partial(m_1-m_3)}{\partial p_2}=-\frac{1}{15}
\neq
\frac{1}{15}=\frac{\partial(m_2-m_3)}{\partial p_1}.
\label{eq:sage_nonconservative}
\end{equation}
Hence the finite-group update is generically non-conservative once the reward distribution has three or more levels. Binary rewards are special because their state is one-dimensional, where the scalar transform $h_{M,\tau}(p)$ exists automatically.

At any finite $M$, the sampled weights still have a useful variational characterization. Let $u_i=1/M$. Then
\begin{equation}
q^\star
=
\arg\max_{q\in\Delta^M}
\left\{
\sum_{i=1}^M q_i R_i
-
\tau\,\mathrm{KL}(q\|u)
\right\},
\qquad q_i^\star=w_i.
\label{eq:sage_variational}
\end{equation}
This standard exponential tilt connects \method{} to RAML, softmax policy gradient, and MPO~\citep{norouzi2016raml,ding2017softmaxpg,abdolmaleki2018mpo}. As $\tau\to\infty$, $A_i=(R_i-\bar R)/\tau+O(\tau^{-2})$, recovering centered reward-weighted policy gradient up to scale. As $\tau\downarrow0$, mass concentrates on the highest-reward rollout(s), yielding a best-of-group update.

\section{Practical PPO optimization}
\label{sec:optimization}

Sections~\ref{subsec:sage_weight_function}--\ref{subsec:sage_general_rewards} established an exact finite-$M$ transform for binary rewards, an exact large-group objective for bounded scalar rewards, and the obstruction to a universal finite-group scalar objective beyond the binary case. None of these results accounts for the off-policy corrections required when optimizing over multiple gradient steps on a fixed rollout batch. This section describes the PPO approximation used in all experiments.

In practice, rollout-level rewards are treated as stop-gradient quantities: we backpropagate through token log-probabilities, but not through reward computation or sampling decisions. We compute rollout-level \method{} advantages once on a batch sampled from $\pi_{\theta_{\mathrm{old}}}$ and optimize a PPO-style clipped surrogate. For rollout $z_i=(z_{i,1},\dots,z_{i,T_i})$, define
\[
\rho_{i,t}(\theta)
=
\frac{\pi_\theta(z_{i,t}\mid x,z_{i,<t})}
{\pi_{\theta_{\mathrm{old}}}(z_{i,t}\mid x,z_{i,<t})}.
\]
We minimize
\begin{equation}
\begin{aligned}
\mathcal L_{\mathrm{\method{}}}^{\mathrm{clip}}(\theta)
&= -
\mathbb E_{\substack{x\sim \mathcal D\\ \mathcal G(x)\sim \pi_{\theta_{\mathrm{old}}}}}
\Bigg[ \\
&\quad
\frac{1}{\sum_{i=1}^M T_i}
\sum_{i=1}^M \sum_{t=1}^{T_i}
\min\Big(
\rho_{i,t}(\theta)A_i,\,
\operatorname{clip}(\rho_{i,t}(\theta),1-\epsilon,1+\epsilon)A_i
\Big)
\Bigg],
\end{aligned}
\label{eq:sage_clip}
\end{equation}
together with a reference-model KL penalty,
\begin{equation}
\begin{aligned}
\mathcal L_{\mathrm{\method{}}}(\theta)
&=
\mathcal L_{\mathrm{\method{}}}^{\mathrm{clip}}(\theta)
\\
&\quad+
\beta\,
\mathbb E_{\substack{x\sim \mathcal D\\ \mathcal G(x)\sim \pi_{\theta_{\mathrm{old}}}}}
\left[
\frac{1}{\sum_{i=1}^M T_i}
\sum_{i=1}^M \sum_{t=1}^{T_i}
\mathrm{KL}\!\Big(
\pi_\theta(\cdot\mid x,z_{i,<t})
\,\|\, 
\pi_{\mathrm{ref}}(\cdot\mid x,z_{i,<t})
\Big)
\right].
\end{aligned}
\label{eq:sage_total}
\end{equation}
This PPO layer should be viewed as a trust-region approximation to the unclipped on-policy objective, not as part of the exact theorem. At $\theta=\theta_{\mathrm{old}}$, clipping is inactive to first order and the leading update direction matches the \method{} estimator; away from that local regime, clipping and reference KL trade objective fidelity for stability.

\custompara{Temperature and stability.}
The same temperature $\tau$ controls both objective geometry and optimizer sharpness. Lower $\tau$ moves the population objective toward MaxRL/ML-like weighting, but it also concentrates the within-group softmax on the highest-reward samples. In practice, smaller $\tau$ therefore requires a tighter trust region---typically a larger reference-KL coefficient, a smaller clip range, or fewer inner-loop updates. We treat this as an optimization issue rather than a change to the underlying population objective.

\custompara{Scope of claims.}
The claims have three distinct scopes. \textbf{Finite-group theorem:} for binary rewards under on-policy unclipped optimization, \method{} induces $h_{M,\tau}(p)$ with derivative $\omega_{M,\tau}(p)$. \textbf{General-reward theorem and limitation:} for bounded scalar rewards, Eq.~\eqref{eq:sage_general_objective} is exact as $M\to\infty$, while Eq.~\eqref{eq:sage_nonconservative} rules out a universal finite-$M$ scalar objective without additional reward assumptions. \textbf{Implementation:} Eqs.~\eqref{eq:sage_clip}--\eqref{eq:sage_total} define the PPO-clipped, reference-KL-regularized approximation used in experiments. Clipping is inactive to first order at $\theta=\theta_{\mathrm{old}}$; away from that local regime, clipping and reference KL trade objective fidelity for stability.

%% file: sections/4_experiments.tex
\section{Experiment Settings}
\label{sec:experiments}

\begin{table*}[t]
 \renewcommand{\grayfont}{}
  \centering
  \renewcommand{\arraystretch}{1.1}
  \setlength{\tabcolsep}{6pt}
  \begin{tabular}{l c c c}
    \toprule
    \multirow{2}{*}{\textbf{Method}} & \multicolumn{3}{c}{\textbf{Benchmarks}} \\
    \cmidrule(lr){2-4}
    & \textbf{GSM8K}
    & \textbf{Countdown}
    & \textbf{DeepMath} \\
    \midrule
    \grayfont Base              & \grayfont 23.0 & \grayfont 2.0  & \grayfont 30.0 \\
    \grayfont SFT               & \grayfont 68.3 & \grayfont 40.7 & \grayfont 35.7 \\
    \grayfont Rationalization   & \grayfont 65.2 & \grayfont 12.5 & \grayfont 34.5 \\
    \grayfont Iterative DPO     & \grayfont 73.1 & \grayfont 40.4 & \grayfont 33.0 \\
    \grayfont RL-Logit          & \grayfont 71.2 & \grayfont 2.2  & \grayfont 37.7 \\
    \grayfont RARO              & \grayfont --   & \grayfont 54.4 & \grayfont 41.3 \\
    \grayfont OPD               & \grayfont \textbf{76.0} & \grayfont 3.4  & \grayfont 42.2 \\
    \grayfont GRPO-Sim           & \grayfont 64.0 & \grayfont 45.1 & \grayfont 38.5 \\
    \grayfont GRPO-Exact         & \grayfont 73.5 & \grayfont 57.7 & \grayfont 50.9 \\
    \midrule
    \method{}-Sim (Ours)         & 71.0 & 48.4 & 39.7 \\
    \method{}-Exact (Ours)       & 75.8 & \textbf{58.1} & \textbf{51.8} \\
    \bottomrule
  \end{tabular}
  \caption{Verifiable reasoning accuracy (\%). ``Sim'' methods use the same weak similarity reward; ``Exact'' methods use the task verifier. Top scores per column are in bold face.}
  \label{tab:rarl_results}
\end{table*}

\subsection{Tasks \& Datasets}
\label{sec:exp:tasks}

We evaluate \method{} across eight benchmarks spanning verifiable and non-verifiable reasoning; all tasks
use a unified \emph{think-then-answer} format with official splits where available (full details in
Appendix~\ref{app:task_details}).
On the \emph{verifiable} side, \textbf{GSM8K}~\citep{cobbe2021training} tests multi-step arithmetic on
grade-school word problems with exact-match accuracy; \textbf{Countdown} is a controlled combinatorial
task requiring four integers to be combined into a target value via basic arithmetic, with correctness
checked by deterministic expression evaluation; and \textbf{DeepMath}~\citep{he2025deepmath} covers
general math reasoning, where answer verification is itself nontrivial.
For \emph{non-verifiable} tasks, \textbf{Poetry Writing} is a custom dataset pairing creative prompts
with expert reference poems, evaluated by an LLM judge~\citep{yang2025qwen3}; and
\textbf{MeetingBank}~\citep{hu-etal-2023-meetingbank} is a long-context summarization benchmark of city
council meeting transcripts, assessed via LLM-as-a-judge.
We additionally report transfer to three standard capability benchmarks from a
separate \textbf{OpenThoughts3-1.2M}~\citep{guha2025openthoughtsdatarecipesreasoning}
training run:
\textbf{AlpacaEval~2.0}~\citep{dubois2024length} (instruction following, length-controlled win rate),
\textbf{MMLU}~\citep{hendrycks2021measuring} (broad academic knowledge),
and \textbf{GPQA}~\citep{rein2024gpqa} (graduate-level science reasoning).

\begin{figure}[t!]
    \centering
    \includegraphics[width=\textwidth]{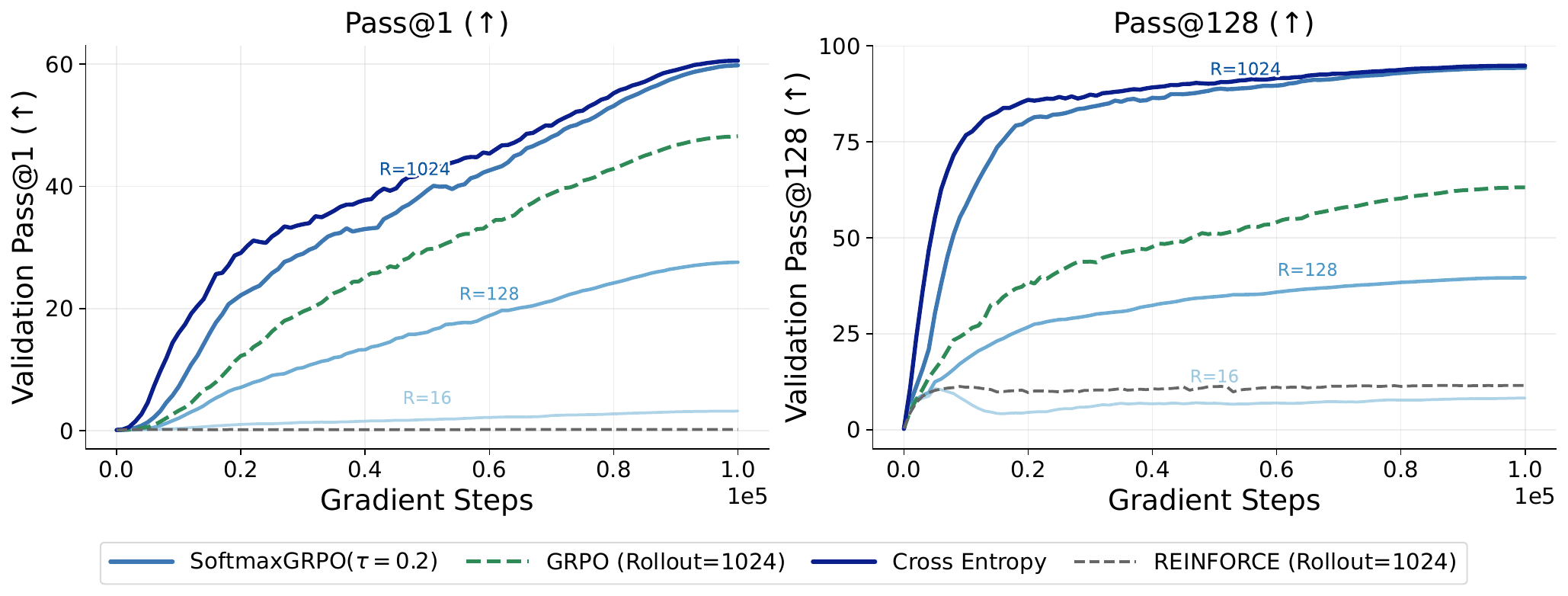}
        \caption{Comparison of training dynamics in ImageNet classification under exact maximum likelihood, REINFORCE, GRPO, and \method{}. With sufficient rollouts, \method{} closely matches cross-entropy training, while REINFORCE fails to make progress from low initial success rates even at large rollout counts.}
    \label{fig:imagenet_results}
\end{figure}

\subsection{Comparisons with Prior Work}
\label{sec:baselines}

We compare \method{} against several post-training methods under identical dataset, training, and
evaluation conditions. Full implementation details are in Appendix~\ref{app:baseline_details}.
\textbf{SFT} maximizes log-likelihood on expert answers directly.
\textbf{Rationalization}~\citep{zelikman2022star} augments each expert answer with a model-generated
chain-of-thought rationale before fine-tuning.
\textbf{Iterative DPO}~\citep{rafailov2023direct,pang2024iterative} runs 3 rounds of preference
optimization from the SFT checkpoint, using on-policy samples as negatives.
\textbf{RL-Logit}~\citep{zhou2025reinforcing,gurung2025learning} trains with rewards derived from the
model's own logits on the expert answer (see Appendix~\ref{app:rl_logit_details}).
\textbf{GRPO-Exact}~\citep{shao2024deepseekmath} applies GRPO with ground-truth binary verifier rewards
on all three verifiable tasks.
\textbf{GRPO-Sim} applies the same GRPO objective with the similarity-based reward $r_{\mathrm{sim}}$
(Appendix~\ref{app:rewards}) in place of a ground-truth verifier, while using standard group
normalization; it therefore tests GRPO under weak rewards on both verifiable and non-verifiable tasks.
Full details are in Appendix~\ref{app:baseline_details}.
\textbf{OPD}~\citep{lu2025onpolicydistillation,agarwal2024policy} distills from a
\texttt{Qwen3-8B}~\citep{yang2025qwen3} teacher via per-token reverse-KL along on-policy trajectories.
\textbf{RARO}~\citep{cai2025escaping} trains a relativistic reasoning critic that provides shaped rewards
for joint policy-critic optimization via GRPO.

\custompara{Training setup.}
Unless otherwise noted, the main experiments fine-tune \texttt{Qwen2.5-1.5B}~\citep{qwen2024qwen2} using
AdamW~\citep{loshchilov2017decoupled} with a learning rate of $1\times10^{-6}$ and bfloat16 precision,
running on NVIDIA H200 and A100 GPUs via the VeRL framework~\citep{sheng2024hybridflow}.
The main GSM8K and Countdown results use $M{=}8$ and $\tau{=}0.1$ for \method{}; DeepMath uses $M{=}16$ and $\tau{=}0.3$, and non-verifiable tasks use $M{=}8$ and $\tau{=}0.3$.
Within every objective-isolation comparison, methods share the model, data, reward, rollout group size, PPO clip, KL coefficient, learning rate, and training budget; only the group-advantage computation differs.
\method{} uses the task's native ground-truth verifier as the reward signal for verifiable tasks (GSM8K, Countdown, DeepMath) and the lightweight similarity-based reward $r_{\mathrm{sim}}$ (Appendix~\ref{app:rewards}) for non-verifiable tasks.
Full training details and per-task reward formulas are in Appendix~\ref{app:training_details}.

\section{Main Results}
\label{sec:main_results}

We organize our main results by reward type: verifiable tasks with programmatic correctness checks (\S\ref{sec:verifiable}), and non-verifiable tasks that require LLM-based or heuristic evaluation (\S\ref{sec:nonverifiable}).

\subsection{ImageNet classification}
\label{sec:imagenet}

ImageNet classification provides a controlled test of how closely \method{} approximates \emph{exact} maximum likelihood in a setting where the latter is available in closed form as the standard cross-entropy objective. We therefore compare four training objectives on ImageNet~\citep{deng2009imagenet} using a ResNet-50~\citep{he2015deepresiduallearningimage}: REINFORCE with a standard baseline, GRPO, \method{}, and exact maximum likelihood. For the RL-style objectives, each rollout samples a class prediction and receives reward $1$ if the predicted class matches the ground-truth label and $0$ otherwise. Full experimental details are provided in the Appendix.

Figure~\ref{fig:imagenet_results} shows a clear gap between expected-reward optimization and maximum-likelihood-style training. REINFORCE fails to make meaningful progress even at large per-example rollout budgets, reflecting the difficulty of learning from sparse binary feedback when initial success rates are low. Exact maximum likelihood, by contrast, exhibits the expected cross-entropy training dynamics. \method{} is trained on the same sampled rollouts and observes the same sparse set of successful trajectories as REINFORCE, but converts this limited signal into a substantially stronger update: as the rollout count increases, it improves steadily and closely tracks exact maximum likelihood. GRPO improves over REINFORCE, but remains visibly farther from the exact maximum-likelihood baseline.

\subsection{Verifiable tasks}
\label{sec:verifiable}

Table~\ref{tab:rarl_results} reports both verifier-based training and a direct objective-isolation comparison under the same weak similarity reward. \method{}-Sim improves over GRPO-Sim on all three tasks: $+7.0$ points on GSM8K, $+3.3$ on Countdown, and $+1.2$ on DeepMath. It also exceeds demonstration-based SFT on GSM8K and Countdown despite using only weak output-overlap rewards. With exact verifier rewards, \method{}-Exact reaches 75.8\% on GSM8K, 58.1\% on Countdown, and 51.8\% on DeepMath. It exceeds GRPO-Exact on all three tasks and attains the best result on Countdown and DeepMath; on GSM8K, it is competitive with OPD (75.8 vs.\ 76.0), which uses dense per-token distillation from a stronger teacher. Appendix~\ref{app:scaling} confirms the advantage at 3B scale.

\subsection{Gradient allocation by prompt difficulty}
\label{sec:gradient_allocation}

\begin{table*}[t]
  \centering
  \small
  \setlength{\tabcolsep}{7pt}
  \renewcommand{\arraystretch}{1.1}
  \begin{tabular}{l l c c c c c}
    \toprule
    \textbf{Task} & \textbf{Method}
    & $[0,0.2)$ & $[0.2,0.5)$ & $[0.5,0.7)$ & $[0.7,0.9)$ & $[0.9,1]$ \\
    \midrule
    \multirow{2}{*}{GSM8K}
      & GRPO      & 4.7\%  & 4.3\%  & 40.1\% & 14.5\% & 36.4\% \\
      & \method{} & 7.3\%  & 33.9\% & 11.9\% & 36.9\% & 10.0\% \\
    \midrule
    \multirow{2}{*}{Countdown}
      & GRPO      & 12.2\% & 20.2\% & 39.6\% & 18.4\% & 9.6\% \\
      & \method{} & 16.0\% & 30.7\% & 33.1\% & 15.1\% & 5.1\% \\
    \bottomrule
  \end{tabular}
  \caption{\textbf{Gradient allocation across prompt difficulty.} Fraction (\%) of total token-level gradient budget assigned by each method to prompts binned by measured pass rate $p$; each row sums to 100\%. \method{} consistently reduces allocation to near-solved prompts ($p\ge0.9$) and shifts budget toward lower-pass-rate examples, most sharply on GSM8K (10.0\% vs.\ 36.4\% for GRPO in the near-solved bin).}
  \label{tab:gradient_allocation}
\end{table*}

The divergence of GRPO's population weight as $p\to1$ does not by itself establish wasted computation, because $\nabla p$ can simultaneously vanish. Table~\ref{tab:gradient_allocation} measures the realized gradient allocation directly. On GSM8K, GRPO spends 36.4\% of its gradient budget on near-solved prompts with $p\ge0.9$, compared with 10.0\% for \method{}. Conversely, \method{} allocates 82.7\% to the moderate-difficulty range $p\in[0.2,0.9)$, compared with 58.9\% for GRPO. Countdown has fewer near-solved prompts, but shows the same shift: \method{} assigns more budget to $p<0.5$ (46.7\% vs.\ 32.4\%) and less to $p\ge0.9$ (5.1\% vs.\ 9.6\%). These measurements support the predicted reallocation away from already-easy prompts and toward examples with greater room to improve.

\subsection{Non-verifiable tasks}
\label{sec:nonverifiable}

\begin{table*}[t]
  \renewcommand{\grayfont}{}
  \centering
  \normalsize
  \renewcommand{\arraystretch}{1.1}
  \setlength{\tabcolsep}{7pt}
  \begin{tabular}{l c c c c c}
    \toprule
    \multirow{3}{*}{\textbf{Method}} & \multicolumn{5}{c}{\textbf{Benchmarks}} \\
    \cmidrule(lr){2-6}
    & \textbf{\shortstack{Poetry\\Score}}
    & \textbf{\shortstack{MeetingBank\\Summ.}}
    & \textbf{\shortstack{AlpacaEval\\2.0}}
    & \textbf{MMLU}
    & \textbf{GPQA} \\
    \midrule
    \grayfont Base              & \grayfont 35.0 & \grayfont 35 & \grayfont 1.61 & \grayfont 60.9 & \grayfont 24.2 \\
    \grayfont SFT               & \grayfont 53.7 & \grayfont 55 & \grayfont 2.18 & \grayfont 61.4 & \grayfont 25.6 \\
    \grayfont GRPO-Sim   & \grayfont 54.6 & \grayfont 62 & \grayfont 2.24 & \grayfont 62.2 & \grayfont 23.8 \\
    \grayfont OPD               & \grayfont 42.6 & \grayfont 42 & \grayfont 2.41 & \grayfont 64.1 & \grayfont 25.3 \\
    \midrule
    \method{} (Ours)               & \textbf{68.0} & \textbf{70} & \textbf{2.50} & \textbf{65.2} & \textbf{27.1} \\
    \bottomrule
  \end{tabular}
  \caption{Non-verifiable task results. We report \textbf{Poetry} score, \textbf{MeetingBank} summarization score (\%), \textbf{AlpacaEval 2.0} length-controlled win rate, \textbf{MMLU}, and \textbf{GPQA} accuracy. Top scores per column are in bold face.}
  \label{tab:nonverifiable}
\end{table*}

Table~\ref{tab:nonverifiable} reports results on five non-verifiable benchmarks spanning creative generation (Poetry), long-context summarization (MeetingBank), instruction following (AlpacaEval~2.0), and general knowledge (MMLU, GPQA). \method{} achieves the best performance across all five tasks, demonstrating that reward-augmented distillation transfers effectively beyond the verifiable regime. The largest gains appear on creative and generative tasks: on Poetry, \method{} scores 68.0, a +13.4 improvement over the next-best baseline (GRPO-Sim, 54.6), and on MeetingBank summarization it reaches 70\% vs.\ 62\% for GRPO-Sim, confirming that the method scales to longer-form generation where programmatic verification is unavailable. \method{} also leads on AlpacaEval~2.0 (2.50 vs.\ 2.41 for OPD), MMLU (65.2 vs.\ 64.1), and GPQA (27.1 vs.\ 25.6), indicating gains on general capabilities alongside task-specific improvements. Cross-judge and blind human calibration are reported in Appendix~\ref{app:judge_validation}.

\begin{table*}[t]
\centering
\small
\begin{subtable}[t]{0.48\linewidth}
  \centering
  \setlength{\tabcolsep}{4pt}
  \renewcommand{\arraystretch}{1.15}
  \begin{tabular}{@{}c cc cc@{}}
  \toprule
  & \multicolumn{2}{c}{$M=4$}
  & \multicolumn{2}{c}{$M=8$} \\
  \cmidrule(lr){2-3}\cmidrule(lr){4-5}
  $\tau$ & Pass@1 & Len & Pass@1 & Len \\
  \midrule
  0.1 & 75.4\% & 102.5 & \textbf{75.8\%} & 100.2 \\
  0.3 & 74.6\% & 101.6 & 75.7\% & 102.8 \\
  0.5 & 74.4\% & 102.9 & 75.0\% & 104.0 \\
  1.0 & 46.1\% & 75.2  & 61.1\% & 92.7  \\
  1.3 & 53.5\% & 99.6  & 63.4\% & 96.8  \\
  1.5 & 41.8\% & 72.8  & 50.2\% & 82.7  \\
  10.0& 31.7\% & 74.0  & 52.5\% & 86.8  \\
  \bottomrule
  \end{tabular}
  \subcaption{\textbf{GSM8K.} Pass@1 (\%) and mean answer length (tokens) across $\tau$ and $M$. Performance is stable for $\tau \le 0.5$ and degrades sharply for $\tau \ge 1.0$, as the reward weighting flattens.}
  \label{tab:gsm8k_sweep}
\end{subtable}
\hfill
\begin{subtable}[t]{0.48\linewidth}
  \centering
  \setlength{\tabcolsep}{4pt}
  \renewcommand{\arraystretch}{1.15}
  \begin{tabular}{@{}c cc cc@{}}
  \toprule
  & \multicolumn{2}{c}{$M=4$}
  & \multicolumn{2}{c}{$M=8$} \\
  \cmidrule(lr){2-3}\cmidrule(lr){4-5}
  $\tau$ & Pass@1 & Len & Pass@1 & Len \\
  \midrule
  0.1 & 54.8\% & 74.0  & \textbf{58.1\%} & 57.8 \\
  0.3 & \textbf{57.8\%} & 60.3  & 55.2\% & 134.3 \\
  0.5 & 29.8\% & 134.0 & 55.8\% & 21.7 \\
  1.0 & 50.0\% & 59.6  & 54.6\% & 22.8 \\
  1.3 & 53.8\% & 70.3  & 55.2\% & 68.7 \\
  1.5 & 51.6\% & 64.6  & 54.6\% & 63.1 \\
  10.0& 46.0\% & 71.8  & 45.2\% & 57.7 \\
  \bottomrule
  \end{tabular}
  \subcaption{\textbf{Countdown.} Pass@1 (\%) and mean response length (tokens) across $\tau$ and $M$. Performance peaks at $\tau \le 0.3$ and degrades at higher $\tau$, with some settings showing anomalous lengths.}
  \label{tab:countdown_sweep}
\end{subtable}
\caption{Temperature ($\tau$) and rollout group size ($M$) ablations on GSM8K (left) and Countdown (right). Both tasks favor low $\tau$; performance degrades as $\tau$ increases and the reward weighting flattens toward a uniform average.}
\label{tab:ablations}
\end{table*}

\subsection{Ablations}

We ablate two key hyperparameters: the temperature $\tau$, which controls how sharply rewards are converted into softmax weights $w_i \propto \exp(R_i/\tau)$ (Eq.~\eqref{eq:sage_weights}); and the rollout group size $M$, which determines how many rollouts are sampled per prompt to compute the group advantage. Table~\ref{tab:ablations} reports a sweep over $\tau\in\{0.1,0.3,0.5,1.0,1.3,1.5,10.0\}$ and $M\in\{4,8\}$ on both GSM8K and Countdown.

\textbf{GSM8K.} Performance is robust to both $\tau$ and $M$ at low temperatures: Pass@1 stays within a tight band (74.4--75.8\%) for $\tau \le 0.5$ across both rollout group sizes, with $M{=}8$ providing a marginal improvement over $M{=}4$ (75.8\% vs.\ 75.4\% at $\tau{=}0.1$). Accuracy degrades sharply once $\tau \ge 1.0$---dropping as low as 31.7\% at $\tau{=}10$---as the soft-max weighting flattens toward a uniform average and the per-step training signal weakens. Answer lengths remain stable ($\approx$100 tokens) throughout the low-$\tau$ regime, with no collapsed runs observed.

\textbf{Countdown.} Countdown is more sensitive to $\tau$, with its best results at $\tau{=}0.1$ for $M{=}8$ (58.1\%) and $\tau{=}0.3$ for $M{=}4$ (57.8\%). Several $(\tau,M)$ configurations exhibit inflated ($>$100 tokens) or collapsed ($<$30 tokens) responses. Sharper softmax weights concentrate the update on fewer rollouts and raise variance, whereas flatter weights weaken the advantage signal; either effect can interact with the PPO trust region and the task's short output format to produce length drift.

Both tasks favor low temperatures, with $\tau\le0.3$ providing the best observed accuracy--stability trade-off. This accords with Section~\ref{subsec:sage_weight_function}: low $\tau$ differentiates high-reward rollouts, whereas larger $\tau$ approaches a weak centered-REINFORCE signal. Because concentration also raises variance, $\tau$ should be tuned jointly with group size, PPO clipping, and reference KL. For rewards normalized to $[0,1]$, $\tau\in[0.1,0.3]$ is a reliable starting range rather than a scale-free default.

%% file: sections/5_conclusions.tex
\section{Discussion and Conclusion}
\label{sec:conclusion}

We introduced \method{}, a drop-in replacement for GRPO that uses temperature-scaled softmax group advantages. Under binary rewards in the on-policy unclipped regime, \method{} admits an exact finite-group population objective with bounded prompt weighting and recovers MaxRL as $\tau\downarrow0$ for finite $M$. For bounded scalar rewards, its large-group update exactly optimizes a log-moment-generating-function objective; at finite $M$, however, the update is generally non-conservative once rewards have three or more levels. This separates the contribution from the established use of exponentiated-reward weights in RAML, softmax policy gradient, and MPO: the new result is the finite-group prompt geometry, its limits, and the boundary of where a scalar objective exists.

The experiments support that geometry directly. Under identical weak rewards, \method{}-Sim improves over GRPO-Sim on all three verifiable tasks, while the gradient-allocation measurement shows that GRPO assigns 36.4\% of its GSM8K gradient budget to prompts with $p\ge0.9$, versus 10.0\% for \method{}. Across the broader evaluation, \method{} reaches 51.8\% on DeepMath with verifier rewards and improves Poetry from 35.0\% to 68.0\% using only lightweight similarity rewards. The scope remains important: the finite-$M$ theorem is exact only for binary rewards under on-policy unclipped optimization, the scalar-reward theorem is asymptotic in group size, the practical method uses PPO clipping and reference KL, the main evaluations focus on a 1.5B model, and non-verifiable evaluation depends on imperfect overlap rewards and LLM judges. Broader scale validation, adaptive temperature selection, and richer process-level rewards remain useful directions.

\vspace{0.05in}
\noindent
{\bf Acknowledgments. } We would like to thank the Ken Kennedy Institute, NSF Career Award \#2201710 and the NSF Campus Cyberinfrastructure grant ``CC* Compute: Interactive Data Analysis Platform'' NSF OAC-2019007, and Rice University’s Center for Research Computing (CRC) for their support.

%% file: sections/7_appendix.tex
\appendix
\counterwithin{figure}{section}
\counterwithin{table}{section}
\renewcommand\thefigure{\thesection.\arabic{figure}}
\renewcommand\thetable{\thesection.\arabic{table}}

\section{Exact \method{} Derivation}
\label{app:sage_exact}

This appendix proves the exact finite-$M$ binary-reward statement used in
Eq.~\eqref{eq:sage_exact_weight} and gives the promised closed-form transform
$h_{M,\tau}$. As in the main text, gradients act only on the log-likelihood
terms in Eq.~\eqref{eq:sage_uc}; rewards and sampled rollouts are treated as
stop-gradient quantities. We fix a prompt $x$ and work in the unclipped
on-policy regime.

Let
\[
p := p_\theta(x)
=
\Pr_{z\sim \pi_\theta(\cdot\mid x)}
\bigl[\mathrm{Dec}(z)\in C(x)\bigr],
\qquad
R_i := \mathbf{1}\!\bigl\{\mathrm{Dec}(z_i)\in C(x)\bigr\},
\]
with
\[
z_1,\dots,z_M \overset{\mathrm{i.i.d.}}{\sim} \pi_\theta(\cdot\mid x),
\qquad
S_i := \sum_{j\neq i} R_j,
\qquad
c := e^{1/\tau}.
\]
We will use the identities
\begin{equation}
\mathbb E\!\left[
R_i \nabla_\theta \log \pi_\theta(z_i\mid x)
\right]
=
\nabla_\theta p_\theta(x),
\qquad
\mathbb E\!\left[
(1-R_i)\nabla_\theta \log \pi_\theta(z_i\mid x)
\right]
=
-\nabla_\theta p_\theta(x),
\label{eq:app_sage_score_identities}
\end{equation}
where the second follows from
$\mathbb E[\nabla_\theta \log \pi_\theta(z_i\mid x)]=0$.

Conditioned on $S_i=s$, \method{} assigns
\[
w_i
=
\begin{cases}
w_s^{(0)} = \dfrac{1}{M+s(c-1)}, & R_i=0,\\[8pt]
w_s^{(1)} = \dfrac{c}{M+(s+1)(c-1)}, & R_i=1.
\end{cases}
\]
Hence the centered advantage $A_i=Mw_i-1$ has the Davis--Recht
conditional-linear form~\citep{davis2025objective}
\begin{equation}
A_i
=
(1-R_i)a_{S_i} + R_i b_{S_i},
\qquad
a_s
=
\frac{M}{M+s(c-1)}-1,
\qquad
b_s
=
\frac{Mc}{M+(s+1)(c-1)}-1.
\label{eq:app_sage_conditional_linear}
\end{equation}
Their gap is
\begin{equation}
\begin{aligned}
\Delta_s^{(\tau)}
:=
b_s-a_s
&=
M\!\left(
\frac{c}{M+(s+1)(c-1)}
-
\frac{1}{M+s(c-1)}
\right)\\
&=
\frac{M(c-1)\bigl(M-1+s(c-1)\bigr)}
{\bigl(M+s(c-1)\bigr)\bigl(M+(s+1)(c-1)\bigr)}.
\end{aligned}
\label{eq:app_sage_delta}
\end{equation}
In particular, $\Delta_s^{(\tau)}>0$ for every $s$, so \method{} always upweights
correct rollouts relative to incorrect ones.

\custompara{Proof of Eq.~\eqref{eq:sage_exact_weight}.}
By exchangeability of the $M$ rollouts,
\[
\mathbb E\!\left[
\nabla_\theta \mathcal J_{\mathrm{\method{}}}^{\mathrm{uc}}(\theta;x,\mathcal G)
\right]
=
\mathbb E\!\left[
A_1 \nabla_\theta \log \pi_\theta(z_1\mid x)
\right].
\]
Condition on $S_1=s$. Since $S_1$ depends only on $\{z_j\}_{j\neq 1}$, it is
independent of $z_1$. Using Eq.~\eqref{eq:app_sage_conditional_linear} and the
score identities in Eq.~\eqref{eq:app_sage_score_identities},
\[
\begin{aligned}
\mathbb E\!\left[
A_1 \nabla_\theta \log \pi_\theta(z_1\mid x)\mid S_1=s
\right]
&=
a_s\,
\mathbb E\!\left[
(1-R_1)\nabla_\theta \log \pi_\theta(z_1\mid x)
\right]
+
b_s\,
\mathbb E\!\left[
R_1\nabla_\theta \log \pi_\theta(z_1\mid x)
\right] \\
&=
(b_s-a_s)\,\nabla_\theta p_\theta(x)
=
\Delta_s^{(\tau)}\,\nabla_\theta p_\theta(x).
\end{aligned}
\]
Averaging over $S_1\sim \mathrm{Binomial}(M-1,p)$ gives
\[
\mathbb E\!\left[
\nabla_\theta \mathcal J_{\mathrm{\method{}}}^{\mathrm{uc}}(\theta;x,\mathcal G)
\right]
=
\mathbb E_{S\sim \mathrm{Binomial}(M-1,p)}
\!\left[
\Delta_S^{(\tau)}
\right]
\nabla_\theta p_\theta(x),
\]
which is Eq.~\eqref{eq:sage_exact_weight}.

\custompara{Closed form for $h_{M,\tau}$ and Bernstein form for $\omega_{M,\tau}$.}
Expanding the binomial expectation yields
\begin{equation}
\omega_{M,\tau}(p)
=
\sum_{s=0}^{M-1}
\Delta_s^{(\tau)}
\binom{M-1}{s}
p^s(1-p)^{M-1-s}.
\label{eq:app_sage_bernstein}
\end{equation}
Thus $\omega_{M,\tau}$ is a Bernstein polynomial of degree $M-1$ with
coefficients $\Delta_s^{(\tau)}$. Since
\[
\frac{d}{dp} I_p(s+1,M-s)
=
\frac{p^s(1-p)^{M-1-s}}{B(s+1,M-s)}
=
M\binom{M-1}{s}p^s(1-p)^{M-1-s},
\]
where $I_p(\cdot,\cdot)$ is the regularized incomplete beta function and
$B(\cdot,\cdot)$ is the beta function, the choice
\begin{equation}
h_{M,\tau}(p)
=
\frac{1}{M}
\sum_{s=0}^{M-1}
\Delta_s^{(\tau)} I_p(s+1,M-s),
\qquad
h_{M,\tau}(0)=0,
\label{eq:app_sage_h_closed_form}
\end{equation}
satisfies
\[
h_{M,\tau}'(p)=\omega_{M,\tau}(p).
\]
This is the exact finite-$M$ scalar transform induced by \method{} under binary
rewards. The main-text expression
$\omega_{M,\tau}(p)=\mathbb E_{S\sim \mathrm{Binomial}(M-1,p)}[\Delta_S^{(\tau)}]$
is simply the binomial-expectation form of
Eq.~\eqref{eq:app_sage_bernstein}.

Because $\omega_{M,\tau}$ is a finite Bernstein polynomial, it is continuous
and bounded on $[0,1]$ for every finite $(M,\tau)$. In particular,
\begin{equation}
\omega_{M,\tau}(1)
=
\Delta_{M-1}^{(\tau)}
=
1-\frac{M}{1+(M-1)e^{1/\tau}}
<
\infty,
\label{eq:app_sage_easy_endpoint}
\end{equation}
so \method{} has no easy-prompt singularity at finite group size.

\custompara{Low-temperature finite-$M$ limit.}
Let $\tau\downarrow 0$, so $c=e^{1/\tau}\to\infty$. Then
\[
\lim_{\tau\downarrow 0}\Delta_s^{(\tau)}
=
\begin{cases}
M-1, & s=0,\\[4pt]
\dfrac{M}{s+1}, & s=1,\dots,M-1.
\end{cases}
\]
Substituting into Eq.~\eqref{eq:app_sage_bernstein} gives
\[
\omega_{M,0}(p)
=
(M-1)(1-p)^{M-1}
+
\sum_{s=1}^{M-1}
\frac{M}{s+1}\binom{M-1}{s}
p^s(1-p)^{M-1-s}.
\]
Using
\[
\frac{M}{s+1}\binom{M-1}{s}=\binom{M}{s+1},
\]
we obtain
\[
\begin{aligned}
\omega_{M,0}(p)
&=
\sum_{s=0}^{M-1}
\binom{M}{s+1}p^s(1-p)^{M-1-s}
-
(1-p)^{M-1} \\
&=
\frac{1}{p}\sum_{k=1}^{M}
\binom{M}{k}p^k(1-p)^{M-k}
-
(1-p)^{M-1} \\
&=
\frac{1-(1-p)^M}{p}
-
(1-p)^{M-1} \\
&=
\frac{1-(1-p)^{M-1}}{p},
\end{aligned}
\]
which proves Eq.~\eqref{eq:sage_maxrl}. This is exactly the MaxRL weighting
with truncation $T=M-1$~\citep{tajwar2026maxrl}.

\custompara{Large-$M$ mean-field form.}
The mean-field form in Eq.~\eqref{eq:sage_weight_mf} is recovered by
substituting the mean of $S\sim\mathrm{Binomial}(M-1,p)$ into
Eq.~\eqref{eq:app_sage_delta}.
Setting $s=(M-1)p$ gives
\[
\begin{aligned}
M-1+s(c-1)\big|_{s=(M-1)p}
&=
(M-1)\bigl(1+p(c-1)\bigr),\\
M+s(c-1)\big|_{s=(M-1)p}
&=
M+(M-1)p(c-1).
\end{aligned}
\]
As $M\to\infty$, both $M+(M-1)p(c-1)$ and $M+(M-1)p(c-1)+(c-1)$ are
$\approx M(1+p(c-1))$, so
\[
\omega_\tau(p)
\approx
\Delta_{(M-1)p}^{(\tau)}
=
\frac{M(c-1)(M-1)\bigl(1+p(c-1)\bigr)}
     {\bigl[M\bigl(1+p(c-1)\bigr)\bigr]^2}
\approx
\frac{c-1}{1+p(c-1)}
=
\frac{e^{1/\tau}-1}{1-p+p e^{1/\tau}},
\]
which is Eq.~\eqref{eq:sage_weight_mf}.
Integrating with respect to $p$ then gives
$h_\tau(p)\propto \log(1-p+p e^{1/\tau})$, up to an additive constant and a
positive $p$-independent scaling.

\custompara{High-temperature limit.}
Let $\tau\to\infty$, so $c=e^{1/\tau}=1+1/\tau+O(\tau^{-2})$.
Writing $\epsilon:=c-1\to 0$ and expanding Eq.~\eqref{eq:app_sage_delta}:
\[
\Delta_s^{(\tau)}
=
\frac{M\epsilon\bigl(M-1+s\epsilon\bigr)}
     {(M+s\epsilon)(M+(s+1)\epsilon)}
=
\frac{(M-1)\epsilon}{M}+O(\epsilon^2).
\]
The leading term is $s$-independent, so the binomial weights in
Eq.~\eqref{eq:app_sage_bernstein} sum to $1$ and give
\[
\omega_{M,\tau}(p)
=
\frac{M-1}{M\tau}+O(\tau^{-2}),
\]
a $p$-independent constant.
To connect this to REINFORCE directly, expand $a_s$ and $b_s$
to first order in $\epsilon$:
\[
a_s\approx -\frac{s\epsilon}{M},
\qquad
b_s\approx \frac{(M-s-1)\epsilon}{M}.
\]
When $R_i=0$ the group has $S_i$ successes total, so
$\bar{R}=S_i/M$ and $A_i=a_{S_i}\approx\epsilon(R_i-\bar{R})$.
When $R_i=1$ the group has $S_i+1$ successes total, so
$\bar{R}=(S_i+1)/M$ and $A_i=b_{S_i}\approx\epsilon(R_i-\bar{R})$.
In both cases,
\begin{equation}
A_i\;\approx\;\frac{R_i-\bar{R}}{\tau}+O(\tau^{-2}),
\label{eq:app_sage_hitemp}
\end{equation}
so \method{} recovers centered group-reward (REINFORCE-style) weighting, scaled
by $1/\tau$. The $p$-independent leading weight $\omega_{M,\tau}(p)\approx
(M-1)/(M\tau)$ is consistent with Eq.~\eqref{eq:app_sage_hitemp}: since
$\mathbb{E}[R_i-\bar{R}]^2 = p(1-p) \cdot \frac{M-1}{M}$ for binary rewards,
every prompt difficulty $p$ receives the same $O(\tau^{-1})$ learning signal.

\subsection{General Scalar Rewards}
\label{app:general_scalar}

We give the full large-group argument behind
Eq.~\eqref{eq:sage_general_objective}. Fix a prompt $x$, write
$g_\theta(z)=\nabla_\theta\log\pi_\theta(z\mid x)$, and let
$X(z)=\exp(R(x,z)/\tau)$. Assume that $R$ is a bounded, measurable,
stop-gradient reward and that the policy score has a finite
$(1+\eta)$-moment for some $\eta>0$. The unclipped group update can be
written as
\begin{equation}
G_M
=
\frac{\frac{1}{M}\sum_{i=1}^M X_i g_\theta(z_i)}
     {\frac{1}{M}\sum_{j=1}^M X_j}
-
\frac{1}{M}\sum_{i=1}^M g_\theta(z_i).
\label{eq:app_general_group_update}
\end{equation}
The law of large numbers, boundedness of $X$, and the score identity give
\begin{equation}
\begin{aligned}
\lim_{M\to\infty}\mathbb E[G_M]
&=
\frac{\mathbb E_{z\sim\pi_\theta}[X(z)g_\theta(z)]}
     {\mathbb E_{z\sim\pi_\theta}[X(z)]}
-\mathbb E_{z\sim\pi_\theta}[g_\theta(z)] \\
&=
\frac{\nabla_\theta Z_\tau(\theta;x)}{Z_\tau(\theta;x)}
=\nabla_\theta\log Z_\tau(\theta;x),
\end{aligned}
\label{eq:app_general_limit}
\end{equation}
where $Z_\tau(\theta;x)=\mathbb E_{\pi_\theta}[e^{R/\tau}]$.
Thus, for bounded scalar rewards, the large-$M$ population update has an
exact scalar objective: the log moment-generating function of reward. The
boundedness assumption can be replaced by the corresponding exponential-
moment and uniform-integrability conditions.

\custompara{Why a universal finite-$M$ objective does not exist.}
For completeness, we expand the counterexample summarized in
Eq.~\eqref{eq:sage_nonconservative}. Let $M=2$ and let the reward take three
values with probabilities $(p_1,p_2,p_3)$ and exponentiated values
$t_k=e^{r_k/\tau}$. Conditional on the first rollout having level $k$, its
expected nonconstant coefficient is
\[
m_k(p)=2\sum_{\ell=1}^3 p_\ell\frac{t_k}{t_k+t_\ell}.
\]
The common centered baseline does not affect integrability. On the simplex
$p_3=1-p_1-p_2$, the update therefore corresponds to the one-form
$(m_1-m_3)\,dp_1+(m_2-m_3)\,dp_2$. For $t=(1,2,4)$, direct substitution gives
\[
\begin{aligned}
m_1&=\frac{3}{5}p_1+\frac{4}{15}p_2+\frac{2}{5},
&m_2&=\frac{2}{3}p_1+\frac{1}{3}p_2+\frac{2}{3},\\
m_3&=\frac{3}{5}p_1+\frac{1}{3}p_2+1,
\end{aligned}
\]
and hence
\[
m_1-m_3=-\frac{1}{15}p_2-\frac{3}{5},
\qquad
m_2-m_3=\frac{1}{15}p_1-\frac{1}{3}.
\]
The cross-partials are $-1/15$ and $1/15$, respectively, so the one-form is
not closed. This rules out a universal finite-group scalar potential once
three or more reward levels are allowed. Additional assumptions that reduce
the reward geometry to a one-dimensional family can restore integrability;
binary rewards are the canonical example.

\custompara{Gaussian example and reward-scale sensitivity.}
Although Gaussian rewards are unbounded, their exponential moments exist. If
$R_\theta\sim\mathcal N(\mu_\theta,\sigma_\theta^2)$, then
\begin{equation}
\log Z_\tau(\theta)
=\frac{\mu_\theta}{\tau}
+\frac{\sigma_\theta^2}{2\tau^2}.
\label{eq:app_gaussian_objective}
\end{equation}
Up to the positive global factor $1/\tau$, the induced objective is
$\mu_\theta+\sigma_\theta^2/(2\tau)$: exponentiation adds a
temperature-controlled variance bonus. If the variance comes from
policy-independent, homoscedastic Gaussian reward noise, its gradient is zero
and the large-$M$ direction reduces to $\tau^{-1}\nabla_\theta\mu_\theta$;
such noise therefore does not bias the population direction.

The same example makes the dependence on reward scale explicit. For
unnormalized weights $X_i=e^{R_i/\tau}$, the large-group effective sample size
satisfies
\begin{equation}
\frac{\mathrm{ESS}}{M}
\longrightarrow
\frac{(\mathbb E X)^2}{\mathbb E[X^2]}
=\exp\!\left(-\frac{\sigma_\theta^2}{\tau^2}\right).
\label{eq:app_gaussian_ess}
\end{equation}
Consequently, $\tau$ must be calibrated to the reward standard deviation.
Our practical range $\tau\in[0.1,0.3]$ is intended for rewards normalized to
$[0,1]$ and should not be transferred unchanged to unbounded, unnormalized
rewards.

\subsection{Robustness to Noisy Binary Rewards}
\label{app:noisy_binary}

Let $Y_i\in\{0,1\}$ denote the clean success indicator and suppose the
observed reward is $R_i=Y_i+\varepsilon_i$, with
$|\varepsilon_i|\le\delta$. If $w_i^0$ is the clean binary \method{} weight
and $\widetilde w_i$ is the noisy weight, then
\begin{equation}
\widetilde w_i
=
\frac{w_i^0 e^{\varepsilon_i/\tau}}
     {\sum_j w_j^0 e^{\varepsilon_j/\tau}}.
\label{eq:app_noisy_weight}
\end{equation}
Because $e^{\varepsilon_i/\tau}\in
[e^{-\delta/\tau},e^{\delta/\tau}]$, every group obeys the multiplicative
envelope
\begin{equation}
e^{-2\delta/\tau}w_i^0
\le \widetilde w_i
\le e^{2\delta/\tau}w_i^0.
\label{eq:app_noisy_envelope}
\end{equation}
Writing $G_M^0$ and $\widetilde G_M$ for the clean and noisy group updates,
respectively, this also gives
\[
\|\widetilde G_M-G_M^0\|
\le
\bigl(e^{2\delta/\tau}-1\bigr)
\sum_i w_i^0\|g_\theta(z_i)\|.
\]
Thus, for fixed $M$ and integrable policy scores, the population update is
perturbed by $O(\delta/\tau)$. If the noise variables are identically
distributed and conditionally independent of the rollouts given the clean
labels, symmetry preserves the one-dimensional binary form, and this statement
can be written directly as
\begin{equation}
\widetilde\omega_{M,\tau}(p)
=\omega_{M,\tau}(p)+O(\delta/\tau).
\label{eq:app_noisy_omega}
\end{equation}

For i.i.d. zero-mean noise, a sharper expected statement follows from
\[
\widetilde w_i
=w_i^0\left[
1+\frac{\varepsilon_i-\sum_jw_j^0\varepsilon_j}{\tau}
\right]
+O(\delta^2/\tau^2).
\]
The first-order term vanishes after conditioning on the clean group. For
bounded noise with variance $\sigma_\varepsilon^2$ and
$\delta/\tau$ small, the expected perturbation is therefore
$O(\sigma_\varepsilon^2/\tau^2)$, with higher-order terms controlled by
$\delta/\tau$. The relevant sensitivity parameter is the noise-to-temperature
ratio: small verifier noise preserves the binary geometry, whereas the bound
becomes uninformative once the noise scale is comparable to $\tau$.

\subsection{Temperature \texorpdfstring{$\tau$}{tau} as a KL Trust-Region Parameter}
\label{app:tau_kl}

Section~\ref{subsec:sage_general_rewards} showed that the \method{} softmax weights are the unique solution of
\[
\max_{q\in\Delta^M}
\Bigl\{
\textstyle\sum_i q_i r_i
-
\tau\,\mathrm{KL}(q\|u)\Bigr\},
\qquad
u_i=\tfrac{1}{M},
\]
with $q^\star_i=w_i\propto e^{r_i/\tau}$. This variational representation gives $\tau$ a precise meaning: it is the Lagrange multiplier (trust-region budget) controlling how far the tilted within-group target deviates from the uniform on-policy empirical prior. Small $\tau$ permits large KL deviations---concentrating mass on high-reward rollouts---while large $\tau$ forces $q$ to remain close to uniform, smoothing the advantage signal toward centered policy gradient. This perspective reframes $\tau$ not as an arbitrary temperature but as a within-group diversity constraint analogous to the KL regularizer in exponentiated-reward policy search~\citep{abdolmaleki2018mpo}.

\custompara{Adaptive $\tau$ via effective sample size.}
For binary rewards, an explicit closed-form rule can target a desired within-group effective sample size (ESS). Suppose $k$ of the $M$ rollouts are correct. With $c=e^{1/\tau}$, the two distinct softmax weights are
\[
w^+ = \frac{c}{kc+M-k},
\qquad
w^- = \frac{1}{kc+M-k},
\]
and the within-group ESS is
\[
\mathrm{ESS}(k,c)
=
\frac{(kc+M-k)^2}{kc^2+M-k}.
\]
Setting $\mathrm{ESS}=\nu$ for a target $\nu\in(k,M]$ and solving for $c$ yields the closed-form rule
\[
c=
\frac{k(M-k)+\sqrt{k(M-k)\,\nu(M-\nu)}}{k(\nu-k)},
\qquad
\tau=\frac{1}{\log c}.
\]
This rule admits an intuitive interpretation: targeting $\nu=M/2$, for example, ensures the update is never dominated by a single rollout regardless of $k$. The rule degenerates when $k=0$ or $k=M$ (all rewards equal, no signal), but those cases require no adaptation. A full empirical evaluation of adaptive $\tau$ selection is left to future work.

\section{Experimental Details}
\label{app:exp_details}

\subsection{Training Hyperparameters}
\label{app:training_details}

All experiments use the VeRL framework~\citep{sheng2024hybridflow} with AdamW~\citep{loshchilov2017decoupled}
optimisation in bfloat16 precision.
Tables~\ref{tab:hyperparams_verifiable} and~\ref{tab:hyperparams_nonverifiable}
summarise the per-task hyperparameters for the verifiable and non-verifiable
training runs, respectively. Unless otherwise noted, all runs use the shared
optimizer, precision, and learning-rate settings stated above.

\begin{table}[H]
\centering
\footnotesize

\begin{tabular}{l ccc}
\toprule[0.95pt]
 & \texttt{GSM8K} & \texttt{Countdown} & \texttt{DeepMath} \\
\midrule[0.6pt]
\multicolumn{4}{c}{\textit{Training Hyper-Parameters}} \\
\midrule[0.6pt]
Base model        & \multicolumn{3}{c}{\texttt{Qwen/Qwen2.5-1.5B}} \\
Optimizer         & \multicolumn{3}{c}{AdamW} \\
Learning rate     & \multicolumn{3}{c}{$1\times10^{-6}$} \\
\method{} temperature $\tau$  & 0.1     & 0.1     & 0.3 \\
PPO clip $\varepsilon$   & \multicolumn{3}{c}{$[0.20,\;0.28]$} \\
KL regularization $\beta$ & \multicolumn{3}{c}{$10^{-3}$} \\
\midrule[0.6pt]
Rollout group size $M$          & 8       & 8       & 16 \\
Rollout batch size              & 64      & 64      & 512 \\
Mini-batch size (per device)    & 4       & 4       & 8 \\
Total training iterations       & 1{,}000 & 2{,}000 & 3{,}220 \\
Max response tokens             & 256     & 256     & 1{,}024 \\
\midrule[0.6pt]
\multicolumn{4}{c}{\textit{Hardware}} \\
\midrule[0.6pt]
GPU device     & $8{\times}$ A100     & $8{\times}$ H200     & $16{\times}$ H200 \\
Compute setup  & 1 node $\times$ 8    & 1 node $\times$ 8    & 2 nodes $\times$ 8 \\
\bottomrule[0.95pt]
\end{tabular}%
\caption{Training hyperparameters for the main verifiable reasoning results.
The \method{} temperature and rollout group size are task-specific; the optimizer,
learning rate, PPO clip, and KL coefficient are shared.}
\label{tab:hyperparams_verifiable}
\end{table}

\begin{table}[H]
\centering
\footnotesize
\begin{tabular}{l ccc}
\toprule[0.95pt]
 & \texttt{Poetry} & \texttt{MeetingBank} & \texttt{OpenThoughts} \\
\midrule[0.6pt]
\multicolumn{4}{c}{\textit{Training Hyper-Parameters}} \\
\midrule[0.6pt]
Base model                     & \multicolumn{3}{c}{\texttt{Qwen/Qwen2.5-1.5B}} \\
Optimizer                      & \multicolumn{3}{c}{AdamW} \\
Learning rate                  & \multicolumn{3}{c}{$1\times10^{-6}$} \\
\method{} temperature $\tau$        & \multicolumn{3}{c}{$0.3$} \\
PPO clip $\varepsilon$         & \multicolumn{3}{c}{$[0.20,\;0.28]$} \\
KL regularization $\beta$      & \multicolumn{3}{c}{$10^{-3}$} \\
\midrule[0.6pt]
Rollout group size $M$         & 8       & 8       & 8 \\
Rollout batch size             & 64      & 512     & 512 \\
Mini-batch size (per device)   & 4       & 4       & 4 \\
Total training iterations      & 1{,}350 & 3{,}220 & 3{,}220 \\
Max response tokens            & 1{,}024 & 4{,}096 & 6{,}144 \\
\midrule[0.6pt]
\multicolumn{4}{c}{\textit{Hardware}} \\
\midrule[0.6pt]
GPU device                     & $8{\times}$ A100  & $16{\times}$ H200 & $16{\times}$ H200 \\
Compute setup                  & 1 node $\times$ 8 & 2 nodes $\times$ 8 & 2 nodes $\times$ 8 \\
\bottomrule[0.95pt]
\end{tabular}
\caption{Training hyperparameters for non-verifiable runs. The OpenThoughts3-1.2M
checkpoint is used for transfer evaluation on AlpacaEval~2.0, MMLU, and GPQA.
Rows above the second rule are shared across all three tasks.}
\label{tab:hyperparams_nonverifiable}
\end{table}

\custompara{Configuration selection and matched comparisons.}
The main GSM8K and Countdown configurations, $(M,\tau)=(8,0.1)$, are the
best-performing cells in the factorial sweep reported in
Table~\ref{tab:ablations}. DeepMath uses $(M,\tau)=(16,0.3)$, while all
non-verifiable runs use the fixed default $(M,\tau)=(8,0.3)$. We did not run
the full sweep on those tasks. Temperature is specific to \method{} and has no
GRPO counterpart. In every \method{}/GRPO objective-isolation comparison, we
match the model, dataset, reward, rollout group size, PPO clip, KL coefficient,
learning rate, rollout sampling, and training budget; only the group-advantage
computation changes.

\subsection{Model-Scale Evaluation}
\label{app:scaling}

We additionally compare verifier-trained \method{} and GRPO at 3B parameters
on GSM8K and Countdown. Within each model scale, the methods use matched
training conditions and differ only in their group-advantage computation.
\begin{table}[H]
\centering
\small
\setlength{\tabcolsep}{7pt}
\begin{tabular}{l l cc}
\toprule
\textbf{Base model} & \textbf{Method} & \textbf{GSM8K} & \textbf{Countdown} \\
\midrule
\multirow{2}{*}{Qwen2.5-1.5B}
& GRPO       & 73.5 & 57.7 \\
& \method{}  & \textbf{75.8} & \textbf{58.1} \\
\midrule
\multirow{2}{*}{Qwen2.5-3B}
& GRPO       & 80.2 & 50.9 \\
& \method{}  & \textbf{82.3} & \textbf{60.4} \\
\bottomrule
\end{tabular}
\caption{Verifier-based accuracy (\%) across model scales under matched
conditions. Bold marks the better objective within each model size.}
\label{tab:scale_results}
\end{table}

The \method{} advantage persists at 3B: it improves over GRPO by 2.1 points
on GSM8K and 9.5 points on Countdown. The Countdown result should not be read
as a monotonic scaling law from two model sizes. In particular, the GRPO
regression from 1.5B to 3B is consistent with verifier-RL scaling behavior
reported independently by \citet{cai2025escaping}; our result establishes that
\method{} does not exhibit that regression in this matched comparison.

\custompara{Reward formulas (verifiable tasks).}
For GSM8K and DeepMath we use the Math-Verify rule-based answer verifier\footnote{\url{https://github.com/huggingface/Math-Verify}}: reward $= 1$ if
the model's final boxed answer matches the gold answer after standard normalisation, and $0$ otherwise.
Every prompt is appended with the instruction \textit{``Please reason step by step, and put your final
answer within \textbackslash{}boxed\{\}.''} For Countdown, correctness is checked by deterministic
evaluation of the predicted arithmetic expression; no additional prompt instruction is added.

\subsection{Task and Dataset Details}
\label{app:task_details}

\noindent\textbf{GSM8K.}
GSM8K~\citep{cobbe2021training} is a dataset of grade-school math word problems requiring multi-step
arithmetic and careful tracking of intermediate quantities.  Each example pairs a natural-language
question with a short, unambiguous final answer.  We report exact-match accuracy after standard answer
normalization.  Training demonstrations, when used, are drawn from the \texttt{HAD653/gsm8k-cot-120b} dataset,\footnote{\url{https://huggingface.co/datasets/HAD653/gsm8k-cot-120b}} which provides chain-of-thought solutions generated by the \texttt{gpt-oss-120b} model.  During \method{} training, we use the verifiable reward function.

\noindent\textbf{Countdown.}
We use a 24-style variant of Countdown where the goal is to combine four integers to obtain 24 using
basic arithmetic operations and parentheses, with each integer used exactly once.  Correctness is
determined by deterministic execution of the predicted expression.  This controlled setting isolates the
role of exploration and credit assignment, since answer checking is far simpler than solution search.
Training demonstrations, when used, are drawn from the verified split of the Countdown-Task-GOLD dataset.\footnote{\url{https://huggingface.co/datasets/HuggingFaceTB/Countdown-Task-GOLD/viewer/verified_Qwen2.5-7B-Instruct}}. During \method{} training, we use the verifiable reward function.

\noindent\textbf{DeepMath.}
DeepMath~\citep{he2025deepmath} covers general math reasoning problems where answer verification is
itself nontrivial, often requiring solving the problem from scratch or handling nontrivial symbolic
manipulation.  Each DeepMath example already includes three solutions generated by
DeepSeek-R1 and verified to be correct; we use the shortest of the three as the training demonstration. During \method{} training, we use the verifiable reward function.

\noindent\textbf{Poetry Writing.}
Poetry Writing is a custom dataset of prompts (topics and optional stylistic constraints) paired with
expert reference poems.  Source poems are drawn from the \texttt{jnb666/poems} dataset.\footnote{\url{https://huggingface.co/datasets/jnb666/poems}}
Since that dataset contains poems only (without accompanying prompts), we use \texttt{gpt-5-instant} to
generate a plausible instruction for each poem---specifically, asking the model what instruction would
most naturally have produced the given poem---yielding (instruction, poem) training pairs.  Overlap-based
rewards only imperfectly capture quality. We generate poems with the instruction
\textit{``You are a helpful assistant that writes poetry.''} For evaluation we
use \texttt{Qwen3-30B-A3B-Thinking-2507}~\citep{yang2025qwen3} as a judge,
scoring poems both in isolation and via pairwise comparison to the expert
reference with the rubric prompt shown in Figure~\ref{fig:poetry_prompts}.

\noindent\textbf{MeetingBank Summarization.}
MeetingBank~\citep{hu-etal-2023-meetingbank} is a long-context benchmark built from public city council
meeting transcripts.  Training transcripts and reference summaries are taken from the
\texttt{microsoft/MeetingBank-LLMCompressed} dataset.\footnote{\url{https://huggingface.co/datasets/microsoft/MeetingBank-LLMCompressed}}
The goal is to generate concise, informative summaries aligned to professionally
written minutes.  Summarization quality is not programmatically verifiable and involves trade-offs
between coverage, faithfulness, and concision. During training, we track ROUGE
and BLEU only. After training, we evaluate the final checkpoint on the held-out
test set with \texttt{Qwen3-30B-A3B-Thinking-2507}~\citep{yang2025qwen3} using
a MeetingBank-specific rubric prompt that scores informativeness, factuality,
fluency, coherence, and conciseness against the transcript segment and
associated meeting metadata, as shown in
Figure~\ref{fig:meetingbank_prompts}.

\noindent\textbf{OpenThoughts3 transfer run.}
For the transfer results on AlpacaEval~2.0, MMLU, and GPQA, we train a separate
non-verifiable \method{} checkpoint on
OpenThoughts3-1.2M~\citep{guha2025openthoughtsdatarecipesreasoning} and
evaluate that checkpoint on the three downstream benchmarks without further
task-specific fine-tuning.

\noindent\textbf{AlpacaEval 2.0, MMLU, and GPQA.}
These three benchmarks assess generalization of capabilities acquired during non-verifiable training.
AlpacaEval~2.0~\citep{dubois2024length} measures instruction-following quality via length-controlled
pairwise win rate against GPT-4.  MMLU~\citep{hendrycks2021measuring} evaluates broad academic
knowledge across 57 subjects.  GPQA~\citep{rein2024gpqa} probes graduate-level reasoning in science
domains.  No training data from these benchmarks is used; results reflect
transfer from the separate OpenThoughts3-1.2M training run.

\subsection{LLM-Judge Validation}
\label{app:judge_validation}

Our primary evaluator is
\texttt{Qwen3-30B-A3B-Thinking-2507} with the task-specific rubrics reproduced
in Figures~\ref{fig:poetry_prompts} and~\ref{fig:meetingbank_prompts}. We
validate these evaluations in two ways. First, we re-score the same outputs
with the architecturally distinct \texttt{gemma-4-31B-it} judge and compare
method rankings. Second, one human rater blindly scores 30 Poetry and 30
MeetingBank outputs without access to the generating method's identity, and we
compare those ratings with the primary judge.

\begin{table}[H]
\centering
\small
\setlength{\tabcolsep}{8pt}
\begin{tabular}{l cc}
\toprule
\textbf{Evaluation} & \textbf{Poetry} & \textbf{MeetingBank} \\
\midrule
Primary judge score            & 68.0  & 70.0 \\
Second judge score             & 59.2  & 65.4 \\
Cross-judge rank correlation   & 0.854 & 0.945 \\
Human--judge correlation       & 0.74  & 0.89 \\
\bottomrule
\end{tabular}
\caption{LLM-judge validation. Scores use the paper's 0--100 reporting scale;
rank correlation compares method-level orderings across the two judges, and
human--judge correlation compares the blind human ratings with the primary
judge on 30 outputs per task.}
\label{tab:judge_validation}
\end{table}

The second judge is systematically stricter in absolute score, but the two
judges agree strongly on method ranking and preserve the \method{}-versus-
baseline ordering on both tasks. Human agreement is higher for MeetingBank,
where factual coverage provides a more objective anchor, than for Poetry,
where quality is intrinsically more subjective. Because the calibration uses
a single human rater, it measures agreement with that rater rather than
inter-rater variability.

\subsection{Reward Formulas and Normalization}
\label{app:rewards}

For verifiable tasks, the reward definitions are given in the training-details
paragraph above. For non-verifiable tasks, we use lightweight answer-level text
similarity rewards. We define
\[
r_{\mathrm{sim}}
=
0.6\,\mathrm{F1}_{\mathrm{SQuAD}}
+
0.4\,\mathrm{ROUGE}\text{-}L.
\]
For datasets with an explicit output-format requirement, the final reward is
\[
r
=
0.35\,r_{\mathrm{format}}
+
0.65\,r_{\mathrm{sim}}.
\]
For datasets without a required format, we use $r=r_{\mathrm{sim}}$. In our
non-verifiable runs, Poetry Writing and MeetingBank use the similarity reward
alone, while the OpenThoughts3-1.2M run additionally includes the format term
to enforce the required thinking-style output structure.

\subsection{Prompt Templates}
\label{app:prompts}

\begin{figure*}[t!]
\centering
\begin{prompt}{Poetry Evaluation Prompt}
{\color{blue}[Evaluation Role]}
\par
You are an expert poetry critic with deep training in New Criticism,
Formalism, and Reader-Response theory. You evaluate poetry based on craft,
execution, and internal logic, not personal taste or agreement with the
subject matter.
\par\relax
{\color{blue}[Goal]}
\par
Your goal is to provide a rigorous critical evaluation of the poem provided
below.
\par\relax
{\color{blue}[Inputs]}
\par
{\color{red}POEM: \{POEM\_TEXT\}}
\par\relax
{\color{blue}[Scoring Guidelines]}
\par
You will score the poem on 7 dimensions. For every dimension, use a standard
Likert scale of 1-10, where:
\par
1-2: Rudimentary / Fails to execute / Accidental.
\par
3-4: Competent but cliche / Lacks tension / Inconsistent.
\par
5-6: Solid execution / Clear intent / Some distinctiveness.
\par
7-8: Excellent craft / High complexity / Strong command of devices.
\par
9-10: Masterful / Transformative / Exceptional handling of the mode.
\par\relax
{\color{blue}[Rubric]}
\par
\textbf{A) Formal Design \& Lineation.} Does the poem have an intelligible
architecture? Do line breaks, stanza shapes, and white space create meaning or
pacing? If a specific form is used, is it handled with discipline or purposeful
variation?
\par
\textbf{B) Sonic Craft \& Rhythm.} Evaluate the ear of the poem. Look for
rhythm (meter or cadence), rhyme (perfect or slant), alliteration, assonance,
and the texture of consonants. Does the sound support the meaning?
\par
\textbf{C) Imagery \& Figurative Language.} Evaluate the ``seeing'' of the
poem. Are images concrete and sensory? Are metaphors and similes fresh and
coherent, or reliant on mixed metaphors and cliches?
\par
\textbf{D) Complexity \& Tension (Close-Reading Depth).} Does the poem
withstand close scrutiny? Look for ambiguity, paradox, irony, and layers of
meaning. Does the poem avoid being overly simplistic or didactic?
\par
\textbf{E) Diction, Syntax, \& Voice.} Evaluate word choice and sentence
structure. Is the diction precise and compressed? Is the syntax used to control
pacing? Is the speaker's persona distinct and controlled?
\par
\textbf{F) Affective Arc \& Resolution.} Evaluate the movement of the poem. Is
there a turn (volta), a realization, or an emotional shift? Does the ending
feel earned?
\par
\textbf{G) Context \& Coherence.} Does the poem succeed on its own terms? If it
engages with history/identity, does it do so with nuance? Does the poem
maintain internal consistency?
\par\relax
{\color{blue}[Output Instructions]}
\par
1. \textbf{Critical Analysis:} First, think step-by-step to justify your
assessment. You MUST explicitly reference specific lines or devices to support
your scores. The scoring calculation happens here.
\par
2. \textbf{JSON Output:} After the analysis, output the scores in valid JSON
format.
\par\relax
{\color{blue}[JSON Output]}
\par
\begin{lmttfont}
\texttt{\char`\{}\\
\hspace*{1em}"A": 1-10,\\
\hspace*{1em}"B": 1-10,\\
\hspace*{1em}"C": 1-10,\\
\hspace*{1em}"D": 1-10,\\
\hspace*{1em}"E": 1-10,\\
\hspace*{1em}"F": 1-10,\\
\hspace*{1em}"G": 1-10\\
\texttt{\char`\}}
\end{lmttfont}
\end{prompt}
\caption{{\bf Poetry Evaluation Prompt.} We evaluate generated poems with
\texttt{Qwen3-30B-A3B-Thinking-2507} using a seven-dimension rubric scored on a
1-10 Likert scale. The resulting aggregate is normalized to a 0--100 scale for
reporting.}
\label{fig:poetry_prompts}
\end{figure*}

\begin{figure*}[t!]
\centering
\begin{prompt}{MeetingBank Evaluation Prompt}
{\color{blue}[Evaluation Role]}
\par
You are an expert evaluator of city-council meeting summaries with deep
training in meeting discourse analysis, public-record minute writing,
procedural language, and factual consistency checking. You evaluate summaries
in the style of MeetingBank: concise, segment-level summaries aligned to a
specific portion of a city council meeting transcript.
\par\relax
{\color{blue}[Goal]}
\par
You judge summaries based on whether they accurately capture the important
content of the meeting segment, especially the main discussion points,
decisions, motions, votes, amendments, referrals, and action items, while
remaining concise, coherent, and readable. You are NOT evaluating creativity,
elegance, or whether you personally agree with the discussion.
\par\relax
{\color{blue}[Inputs]}
\par
{\color{red}MEETING TITLE (optional): \{MEETING\_TITLE\}}
\par
{\color{red}TRANSCRIPT SEGMENT: \{TRANSCRIPT\_TEXT\}}
\par
{\color{red}CANDIDATE SUMMARY: \{SUMMARY\_TEXT\}}
\par\relax
{\color{blue}[Non-Negotiable Rules]}
\par
Judge ONLY against the transcript segment and provided metadata; do not use
external knowledge. Faithfulness is more important than fluency, and small
factual errors matter, including names, departments, bill numbers,
ordinance/resolution IDs, dates, dollar amounts, motions, amendments, vote
outcomes, referrals, and assigned follow-up actions.
\par
Do NOT reward plausible inventions. If a detail is not in the transcript, it
should not be in the summary. Do NOT penalize a summary for reusing wording
from the transcript when that improves accuracy and concision.
\par
If the segment contains a decision, motion, amendment, vote, referral, or
action item, omission or distortion of that procedural outcome is a major
error. If the segment contains no explicit decision or action item, do NOT
hallucinate one; instead reward summaries that correctly characterize the
segment as discussion, testimony, presentation, clarification, deliberation, or
public comment.
\par
If the transcript itself contains contradictions, do NOT penalize a summary for
faithfully reflecting the ambiguity or for choosing the corrected/final
version; do penalize a summary that resolves genuine ambiguity by inventing a
false certainty. If the segment is purely procedural, brevity is appropriate.
If the segment is largely inaudible, garbled, or mostly crosstalk, reward
summaries that honestly acknowledge the limited content rather than fabricating
substance.
\par\relax
{\color{blue}[Scoring Rubric]}
\par
Score each dimension on a 1-10 Likert scale:
\par
1-2: Fails badly / misleading / unusable / misses the point entirely.
\par
3-4: Weak / major omissions or distortions / low utility.
\par
5-6: Adequate / partially correct / noticeable problems but has some value.
\par
7-8: Strong / accurate / useful / well-compressed.
\par
9-10: Excellent / highly faithful / sharp / near-reference quality.
\par
\textbf{A) Informativeness.} Does the summary capture the main points of the
meeting segment at an appropriate level of compression? A strong summary
contains all and only the important information: the core issue under
discussion, the substance of the debate, the main proposal, the most important
concerns raised, and any meaningful outcome, condensed to a length proportional
to the segment's substantive content.
\par
\textbf{B) Factuality.} Are all facts in the summary consistent with the
transcript segment? Check who said or did what, what was proposed, whether
something was approved, denied, deferred, amended, tabled, or referred, any
vote result, any action item or next step, and all important entities, numbers,
and procedural details.
\par
\textbf{C) Fluency.} Are the sentences clear, grammatical, and readable? A
strong summary should be well-written, precise, and easy to understand by a
reader unfamiliar with the meeting.
\par
\textbf{D) Coherence.} Does the summary fit together as a unified account of
one meeting segment? A strong summary should present information in a logical
order and stay focused on the same agenda item or discussion thread.
\par
\textbf{E) Conciseness.} Does the summary avoid unnecessary repetition and
filler while remaining information-dense? Higher scores mean the summary is
MORE concise and information-dense.
\par\relax
{\color{blue}[Output Instructions]}
\par
\textbf{Step 1 --- Critical Analysis.} Provide an evidence-grounded analysis
that justifies your evaluation. You MUST explicitly reference concrete details
from the transcript segment and the candidate summary, identify what the
summary gets right and wrong, and note any hallucinated content.
\par
\textbf{Step 2 --- Per-Dimension Verdicts.} For each dimension, write one
sentence stating the score and the single most important reason for that score,
formatted as:
\par
\texttt{Informativeness [X/10]: <reason>}
\par
\texttt{Factuality [X/10]: <reason>}
\par
\texttt{Fluency [X/10]: <reason>}
\par
\texttt{Coherence [X/10]: <reason>}
\par
\texttt{Conciseness [X/10]: <reason>}
\par
\textbf{Step 3 --- JSON Scores.} Output the final scores in valid JSON format.
\par\relax
{\color{blue}[JSON Output]}
\par
\begin{lmttfont}
\texttt{\char`\{}\\
\hspace*{1em}"informativeness": 1-10,\\
\hspace*{1em}"factuality": 1-10,\\
\hspace*{1em}"fluency": 1-10,\\
\hspace*{1em}"coherence": 1-10,\\
\hspace*{1em}"redundancy": 1-10\\
\texttt{\char`\}}
\end{lmttfont}
\end{prompt}
\caption{{\bf MeetingBank Evaluation Prompt.} We evaluate held-out MeetingBank
summaries with \texttt{Qwen3-30B-A3B-Thinking-2507} using a five-dimension
rubric over informativeness, factuality, fluency, coherence, and conciseness.
The JSON output records the final conciseness score under the
\texttt{redundancy} field, matching the evaluation pipeline.}
\label{fig:meetingbank_prompts}
\end{figure*}

\subsection{RL-Logit Reward Variants}
\label{app:rl_logit_details}

We implement two variants of logit-based rewards following \citet{zhou2025reinforcing,gurung2025learning}.
\textbf{Log-probability reward:} the scalar reward is $\log \pi_{\theta}(a^{\star} \mid q, z)$, the
log-probability of the expert answer $a^{\star}$ given the question $q$ and the generated reasoning
trace $z$.
\textbf{Perplexity reward:} the reward is the negative perplexity of the expert answer under the same
conditional distribution, $-\exp\!\bigl(-\tfrac{1}{|a^{\star}|}\log \pi_{\theta}(a^{\star} \mid q, z)\bigr)$.
We report the best-performing variant for each task.

\subsection{Baseline Implementation Details}
\label{app:baseline_details}

\noindent\textbf{Rationalization.}
We prompt the base model to annotate each expert demonstration with a free-form rationale, then perform
SFT on the concatenated (question, rationale, answer) sequences.  This baseline is designed to incentivize
the model to produce explicit reasoning before the final answer, following the STaR framework~\citep{zelikman2022star}.

\noindent\textbf{Iterative DPO.}
Inspired by Iterative Reasoning Preference Optimization~\citep{pang2024iterative}, we perform 3 rounds of
DPO iteratively.  In each round, we sample one response per question from the current policy to form
preference pairs favoring the expert answer over the on-policy sample.  We initialize from the SFT
checkpoint to mitigate distribution mismatch and report the best performance across rounds.

\noindent\textbf{On-Policy Distillation (OPD).}
We closely follow the setting described in~\citet{lu2025onpolicydistillation} (see also~\citet{agarwal2024policy}).
We sample on-policy rollouts from the student and, at each visited prefix, query the teacher
(\texttt{Qwen3-8B}~\citep{yang2025qwen3}) to obtain its next-token distribution under the same
student-conditioned context.  The student is then updated to minimize a per-token reverse-KL distillation
loss toward the teacher along these trajectories.  This provides dense token-level supervision while
avoiding the training--inference mismatch of purely off-policy distillation.

\noindent\textbf{RARO.}
RARO~\citep{cai2025escaping} learns a \emph{reasoning critic} that performs a \emph{relativistic} pairwise
comparison between a policy answer and the corresponding expert answer, predicting whether the expert is
better, the policy is better, or whether they are tied.  The critic's prediction induces a shaped reward:
the critic is rewarded for correctly identifying the expert in the pair, while the policy is rewarded for
``fooling'' the critic; the explicit tie option mitigates critic degeneracy near optimality and stabilizes
training.  In practice, RARO jointly optimizes both policy and critic with GRPO, using key stabilizers
such as sharing parameters between the critic and policy, mixing policy/critic rollouts within a batch,
and sampling critic prompts from a replay buffer of past expert and policy answers to reduce catastrophic
forgetting.

\noindent\textbf{GRPO-Sim and \method{}-Sim.}
Both weak-reward variants use the same similarity reward
$r_{\mathrm{sim}}$ defined in Appendix~\ref{app:rewards}. Concretely,
$r_{\mathrm{sim}}=0.6\,\mathrm{F1}_{\mathrm{SQuAD}}+
0.4\,\mathrm{ROUGE}\text{-}L$ measures token-level overlap between the model
output and the reference demonstration; tasks with a required output format
also use the format term described in Appendix~\ref{app:rewards}. GRPO-Sim
applies standard GRPO group normalization~\citep{shao2024deepseekmath}, whereas
\method{}-Sim replaces only that advantage computation with
Eq.~\eqref{eq:sage_weights}. The paired runs match the model, data, reward,
rollout group size, PPO clip, KL coefficient, learning rate, rollout sampling,
and training budget. Consequently, their difference in
Table~\ref{tab:rarl_results} isolates group-advantage geometry under an
identical weak reward. The Exact variants provide the complementary comparison
under each task's binary verifier.